\documentclass{article}

\usepackage{microtype}
\usepackage{graphicx}
\usepackage{subfigure}
\usepackage{booktabs} 
\usepackage{lipsum}

\usepackage{hyperref}

\usepackage[accepted, nohyperref]{icml2024}
\usepackage{hyperref}

\usepackage{amsmath}
\usepackage{amssymb}
\usepackage{mathtools}
\usepackage{amsthm}

\usepackage[capitalize,noabbrev]{cleveref}
\usepackage{mdframed}

\newcommand{\ElvisIssue}[1]{}
\newcommand{\ElvisFix}[1]{}

\theoremstyle{plain}
\newtheorem{theorem}{Theorem}[section]

\newtheorem{lemma}[theorem]{Lemma}
\newtheorem{corollary}[theorem]{Corollary}
\theoremstyle{definition}

\theoremstyle{remark}
\newtheorem{remark}[theorem]{Remark}

\usepackage[textsize=tiny]{todonotes}

\icmltitlerunning{A Tale of Tails: Model Collapse as a Change of Scaling Laws}

\begin{document}

\twocolumn[
\icmltitle{
A Tale of Tails: Model Collapse as a Change of Scaling Laws
}



\icmlsetsymbol{equal}{*}

\begin{icmlauthorlist}
\icmlauthor{Elvis Dohmatob}{equal,fair}
\icmlauthor{Yunzhen Feng}{equal,cds}
\icmlauthor{Pu Yang}{pu}
\icmlauthor{Francois Charton}{fair}
\icmlauthor{Julia Kempe}{cds,courant,fair}
\end{icmlauthorlist}

\icmlaffiliation{cds}{Center for Data Science, New York University}
\icmlaffiliation{courant}{Courant Institute, New York University}
\icmlaffiliation{fair}{Meta FAIR}
\icmlaffiliation{pu}{School of Mathematical Sciences, Peking University}

\icmlcorrespondingauthor{Yunzhen Feng}{yf2231@nyu.edu}

\icmlkeywords{Machine Learning, ICML}

\vskip 0.3in
]



\printAffiliationsAndNotice{\icmlEqualContribution} 

\begin{abstract}
As AI model size grows, neural {\em scaling laws} have become an important tool to predict the improvements of large models when increasing capacity and the size of original (human or natural) training data. Yet, the widespread use of popular models means that the ecosystem of online data and text will co-evolve to progressively contain increased amounts of synthesized data.
In this paper we ask: {\em How will the scaling laws change in the inevitable regime where synthetic data makes its way into the training corpus?}
Will future models, still improve, or be doomed to degenerate up to total {\em (model) collapse}?  We develop a theoretical framework of model collapse through the lens of scaling laws. We discover a wide range of decay phenomena, analyzing loss of scaling, shifted scaling with number of generations,  the ``un-learning" of skills, and grokking when mixing human and synthesized data. Our theory is validated by large-scale experiments with a transformer on an arithmetic task and text generation using the large language model {\tt Llama2}.

\end{abstract}

\section{Introduction}


Groundbreaking advances in generative AI algorithms for text, images and code are ushering in the "synthetic data age": increasingly we consume data generated by large scale models like GPT4 \cite{achiam2023gpt}, Stable Diffusion \cite{Rombach_2022_CVPR} and their successors. 
A growing number of synthetic data generated with these models starts to populate the web, often indistinguishable from ``real" data.  Evidence suggests that AI-generated content has contaminated the LAION-5B dataset \citep{alemohammad2023selfconsuming} and has been utilized by crowdworkers \citep{veselovsky2023artificial}. Remarkably, as of the current assessment, ChatGPT contributes to 0.1\% of the total words generated daily by the global population \citep{daily_generate}. 

At the same time a key driver behind the current success of large models is their consumption of massive amount of web-scale data for training. The improvements of larger models are governed by scaling laws in which error falls off as a power in the size of training data; and the emergence of new skills seems tightly linked to covering increased scales of training data. Our understanding of what the future holds in a world were models are trained on other models (or their own) synthesized data is only at its beginning, but some works indicate the possibility of complete collapse of learning, so called model collapse\footnote{Not to be confused with neural collapse which refers to clustering of last-layer features at the end of training \cite{papyan2020prevalence}}.


\begin{figure}[tb]
\vspace{-0.3cm}
    \centering
    \includegraphics[width=\linewidth]{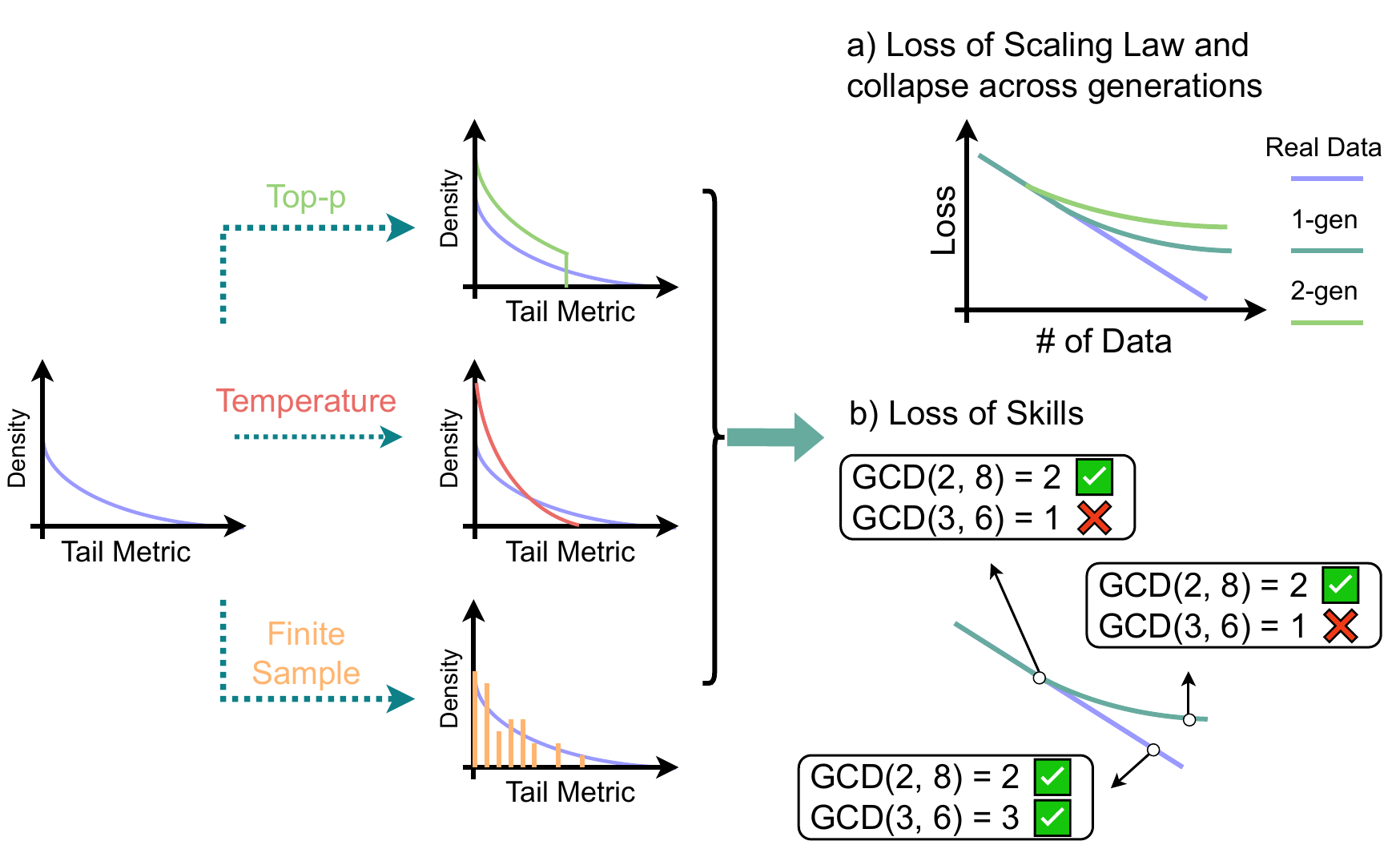}
    \caption{Top-$p^{inf}$ (nucleus) sampling, temperature scaling of LLM generation, and finite sample bias lead to truncated or narrowed ``tails" (left side), causing loss of scaling laws (top right) and loss of skills (bottom right). Here we visualize calculating the greatest common divisor (GCD) with answer 2 and 3 as two skills.}
    \label{fig:teaser}
    \vspace{-0.6cm}
\end{figure}

{\bf Scaling Laws.} In many domains of machine learning including speech, translation, vision, video, and mathematical problem solving, empirically observed neural scaling laws \cite{hestness2017deep,rosenfeld2020a,kaplan2020scaling,hoffmann2022trainingChinchilla,gordon-etal-2021-data,henighan2021scaling,aghajanyan2023scaling}
demonstrate that test error often falls off as a power law with the amount of training data, model size, and compute. Theoretically, scaling laws have been derived in a variety of settings (e.g. \citet{hutter2021learning,cabannes2023scaling} for ``LLM-like" models).

Scaling laws are intimately related to the emergence of abilities \cite{wei2022emergent} in larger models, that are not present in smaller ones; and to skills that appear with decreasing loss \cite{gordon-etal-2021-data,arora2023theory}.
%
This bolsters the now common narrative that ``scaling is all you need". 

\paragraph{Model Collapse.} Current LLMs \cite{devlin2018bert, liu2019roberta, NEURIPS2020_1457c0d6, touvron2023llama}, including GPT-4 \cite{achiam2023gpt}, were trained on predominantly human-generated text; similarly, diffusion models like DALL-E \cite{pmlr-v139-ramesh21a}, Stable
Diffusion \cite{Rombach_2022_CVPR}, Midjourney \cite{midjourney2023} are trained on web-scale image datasets. 
These training corpora already potentially exhaust all the available clean data on the internet. 
A growing number of synthetic data generated with these increasingly popular models starts to populate the web, often indistinguishable from ``real" data.  
 This proportion is expected to grow, leading us to anticipate a future where AI-generated data predominates.
We have thus already entered a future where our training corpora are irreversibly mixed with synthetic data, and this situation is likely to worsen.
Recent works call attention to the potential dramatic deterioration in the resulting models, an effect referred to as {\em ''model collapse"} \cite{shumailov2023curse}. Facets of this phenomenon have been demonstrated {\em empirically} in various settings \cite{lebrun2021evaluating,Hataya_2023_ICCV, martínez2023combining,martínez2023understanding,bohacek2023nepotistically,briesch2023large,guo2023curious}. Theoretically, a few works are emerging to analyze the effect of iterative training on self-generated (or mixed) data: \cite{shumailov2023curse} coin the term {\em``model collapse"} to characterize complete reversion to the mean, \citet{alemohammad2023selfconsuming} analyze {\em ``self-consuming loops"},  \citet{bertrand2023stability} show that iterative synthetic training leads to a {\em ``clueless generator"}, and \citet{dohmatob2024model} analyze n-fold generation with kernel regression.

With these first warning signs in place, we thus ask:

\begin{mdframed}
{\em How is the current scaling paradigm affected by synthetic data in the training corpus?}
\end{mdframed}

\begin{figure}[htb]
    \centering
    \vspace{-5pt}
    \includegraphics[width=0.7\linewidth]{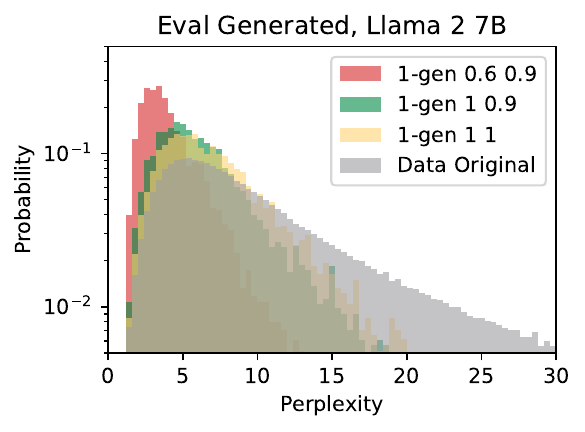}
    \includegraphics[width=0.64\linewidth]{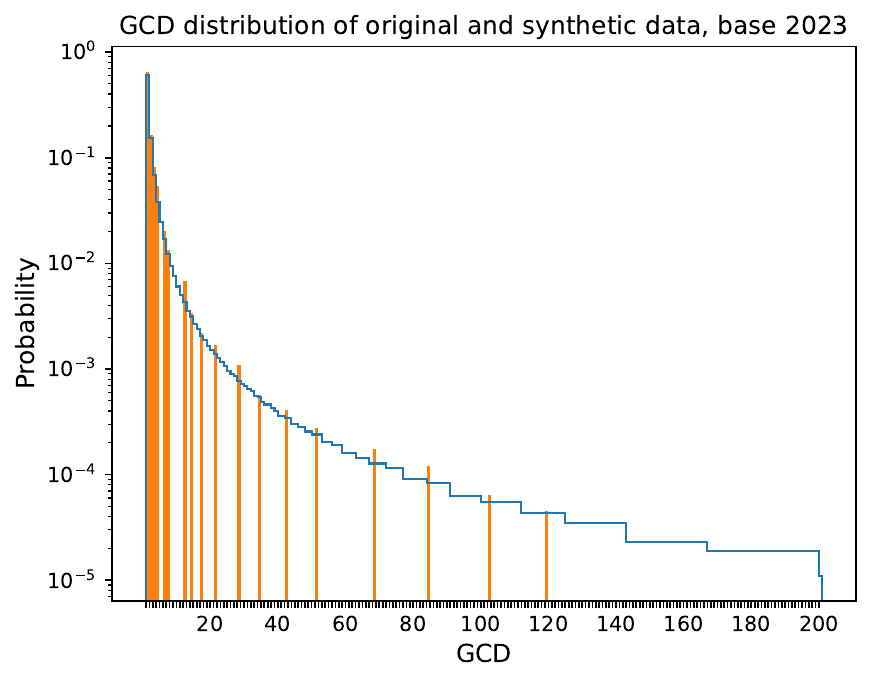}\label{fig:gcd_base2023}
    \vspace{-.4cm}
    \caption{
    {\bf Tails of AI-generated data:} {\bf Top.} Perplexity diagram of the Wikitext-103 test set, measured with {\tt Llama2-7B} as the anchor model. We query the Wikitext-finetuned {\tt Llama2-7B} to generate AI data, which is compared to the original set. Perplexity is calculated solely for the generated positions in both the AI and original datasets. AI data is generated for various settings of $(p^{inf},\tau)$.
    {\bf Bottom.}  Distribution of greatest common divisors (GCD) of pairs of random integers (original data (blue) scaling as $p(GCD=k) \propto k^{-2}$). A transformer is trained on this task on $300M$ samples and used as a generator on a test set of randomly sampled integer pairs, giving the truncated GCD distribution.}
        \vspace{-0.5cm}
    \label{fig:wiki-raw-test-evalgen}
\end{figure}




To this end, we carefully zoom into the scaling behavior of LLM-style models.
Theoretical derivations of scaling laws always assume a heavy-tailed distribution (power-law, aka Zipf) on the input features (``heavy tail in, power scaling law out"). This distribution is of the form
\begin{equation}\label{eq:zipf}
    p_i \propto i^{-\beta},\,\,\,\, i = 1,2,\ldots 
\end{equation}
Such distributions are ubiquitous in natural datasets, from Zipf's law \cite{zipf1935psycho} in distribution of word frequencies, to biological data, earthquake magnitudes, financial data etc. - this is the data being consumed by large models at scale, like LLMs.
But what distribution do AI-models generate when trained on such data? Figure \ref{fig:wiki-raw-test-evalgen} provides an empirical answer for a large scale LLM (Llama2-7B) and a transformer model trained on an arithmetic task. 
Regenerating heavy-tailed data affects the distribution in two possible ways: (1) ``Cutting off" the tail of the distribution and/or (2) ``Narrowing" the tail (see Figure \ref{fig:teaser} for a cartoon illustration). The mechanisms leading to this, apart from finite sampling bias (as already proposed in \citet{shumailov2023curse} - see Section \ref{sec:hutter} for a derivation in the Zipf-setting), stem from deliberate choices in the generation algorithms of the models: in LLMs via truncated next token prediction at inference (e.g. selecting more likely tokens via {\em top-$p^{inf}$} or {\em top-$k^{inf}$} truncation, concentrating the probability distribution by lowering the temperature $\tau$ ); in vision models like GANs via truncation or in diffusion models through guidance.

\begin{figure*}[htb]
    \centering
    \vspace{-6pt}
    \includegraphics[width=.32\textwidth]{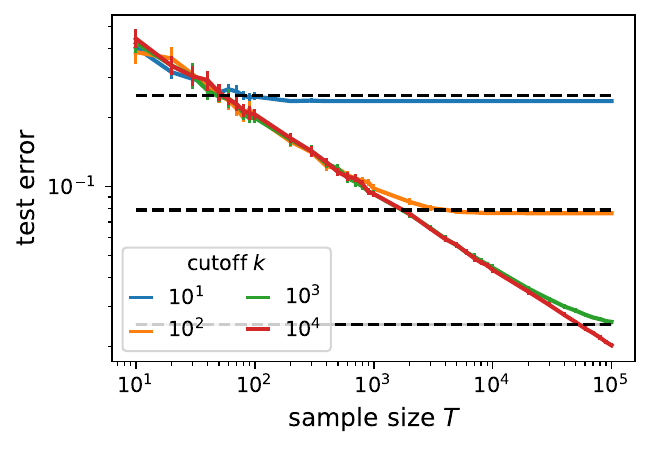}
    \includegraphics[width=0.32\textwidth]{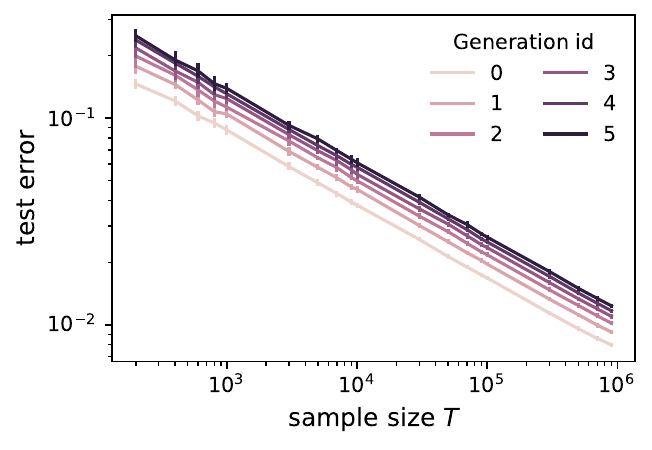}
    \includegraphics[width=.32\textwidth]{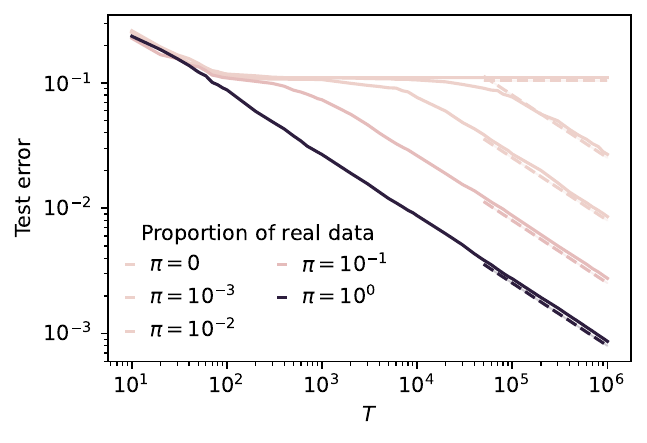}
    \vspace{-.5cm}
    \caption{\textbf{Illustration of Our Main Results for Simplified LLMs.}
    \textbf{Left plot.} Empirical confirmation of the double scaling law. The true distribution of the data is Zipf with exponent $\beta = 3/2$. Broken lines correspond to $k^{-(\beta-1)}$, for varying $T$ and different values of $k$. \textbf{Middle plot.}
    Model collapse over multiple generations. Again $\beta = 3/2$,  $T_0 = T$ across all generations with no additional tail-cutting, regeneration for 5 times.
     \textbf{Right plot.} Notice the grokking behavior, as perfectly predicted by the Theorem \ref{thm:grokk}. For any given value $\pi$ for the proportion of real data, the broken lines are asymptotes $E_{test} \asymp (\pi T)^{-c}$ and each plateau has length of order $k^\beta/\pi$, both predicted by the theorem. See Figure \ref{fig:grokk-old} for similar results with other values of $k$.}
    \label{fig:grokk}
\end{figure*}
\begin{figure*}[htb]
    \centering
    \vspace{-8pt}
    \includegraphics[width=0.3\textwidth]{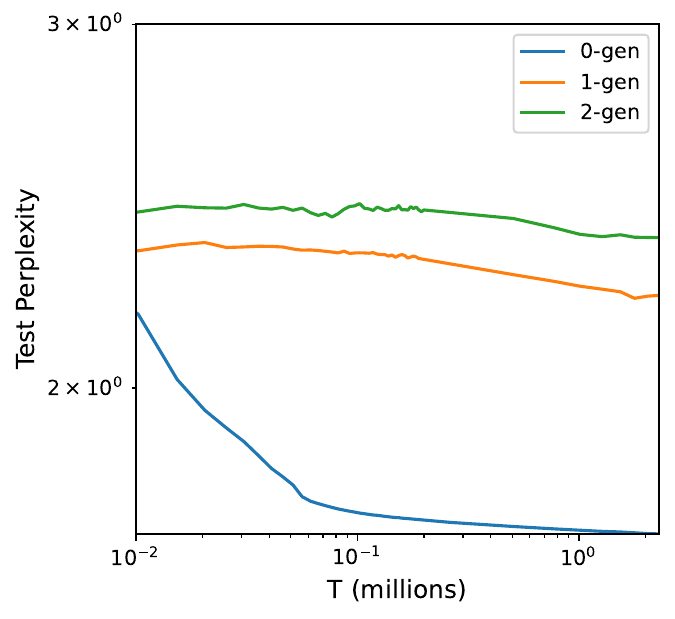}
    \includegraphics[width=0.3\linewidth]{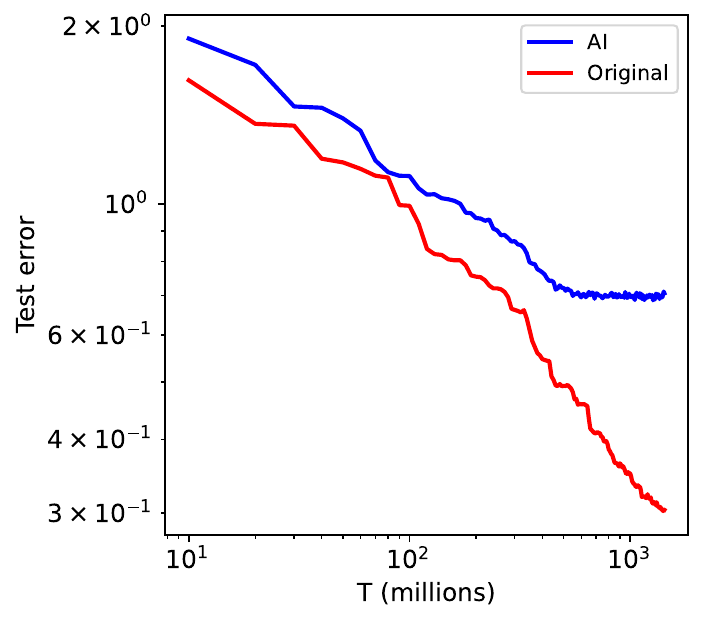}
        \includegraphics[width=0.39\linewidth]{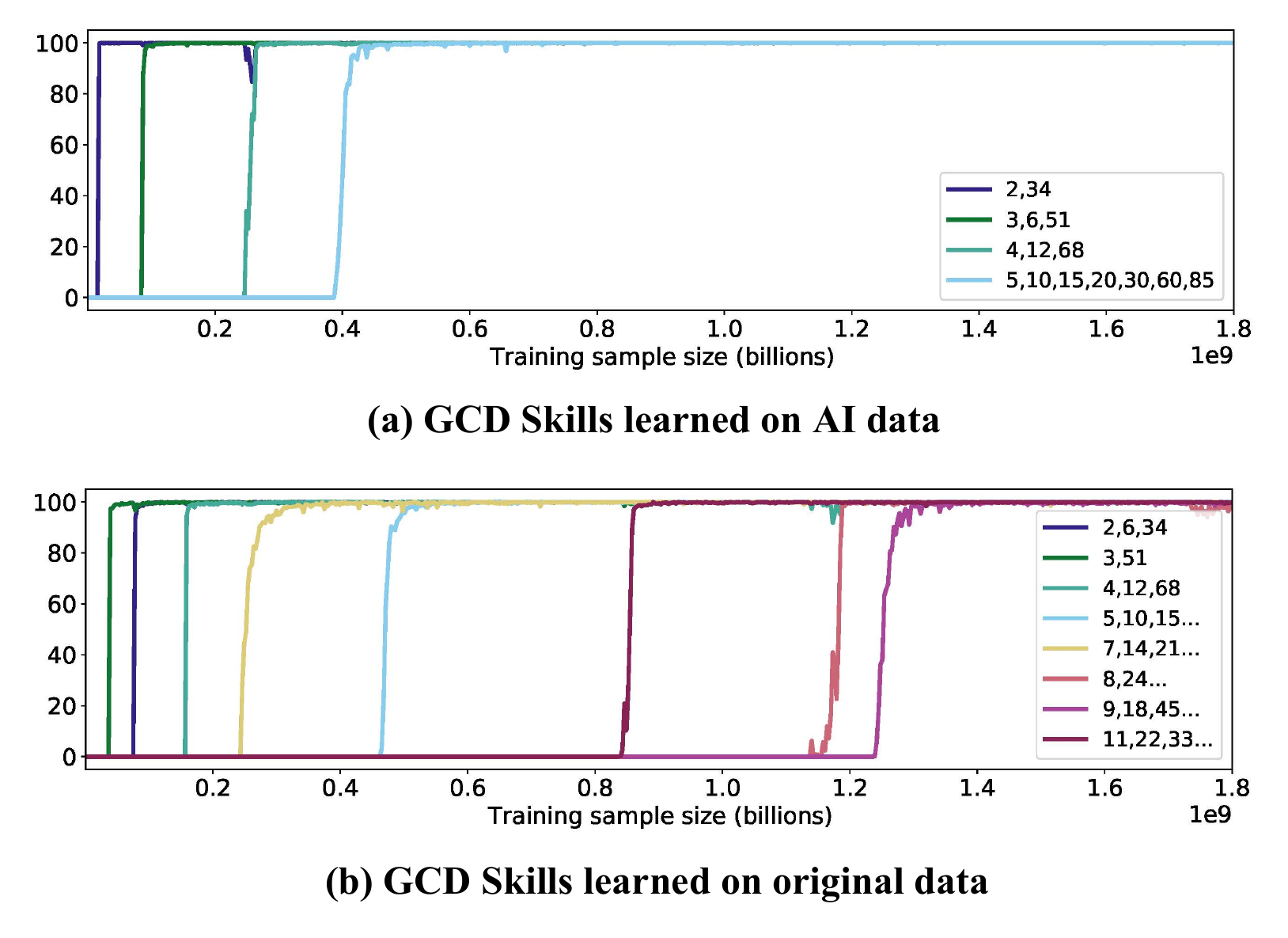}
    \vspace{-.7cm}
    \caption{\textbf{Experimental Results (Details in Section \ref{sec:experiments}).}
    \textbf{Left plot.} The scaling law for finetuning {\tt Llama2-7B} on the Wikitext-103 dataset. '0-gen' utilizes the original data, while subsequent generations use data generated by the previous model. 
    \textbf{Middle plot.} Scaling law of the transformer model trained to predict GCD of two integers. Data is synthesized from a 0th generation model trained on $300K$ samples. Note the tapered-off scaling of the model trained on synthesized data, as predicted by our theory. \textbf{Right plot.} ''Skills" (bursts of new GCDs) learned by the GCD-transformer on original (bottom) and AI data (top). We see how the disappearance of scaling leads to the disappearance of abilities, mastered by the model trained on clean data. }
    \vspace{-0.5cm}
    \label{fig:experiments}
\end{figure*}

{\bf Summary of Main Contributions.}
We present a high-level summary of our main theoretical contributions, some of which are highlighted in Figure \ref{fig:grokk}.
We empirically verify these theoretical predictions (see Figure \ref{fig:experiments}): (1) in large-scale experiments on an LLM, fine-tuning Llama2-7B \cite{touvron2023llama} on an approximately $2M$ sample dataset from Wikitext-103 and (2) for  transformer models trained to predict the greatest common divisor \cite{charton2023transformers}.
   
Assuming a true distribution as in Equation \eqref{eq:zipf}, consider training a model on $T$ AI data-generated data. The synthesized data amounts to a version of the true data distribution with the tail cut at some finite rank $k$ or the tail narrowed to a smaller exponent.
Our main findings are as follows. 


 \textbf{\textit{(1) A Double Scaling Law.}}
 We establish new scaling laws that explain model collapse in simplified (non-autoregressive) LM \cite{hutter2021learning} and toy bigram LLMs 
 (refer to Theorems \ref{thm:simple} and \ref{thm:bigram3})\footnote{The notation $f(T) \lesssim g(T)$ means that $f(T) \le C g(T)$ for sufficiently large $T$ and an absolute constant $C$, while $f(T) \asymp g(T)$ means $f(T) \lesssim g(T) \lesssim f(T)$.}
    \begin{eqnarray}
    E_{test} \asymp T^{-c} + k^{-c'}. 
    \label{eq:plateau1}
    \end{eqnarray}
    or equivalently (refer to Corollary \ref{cor:from-k-to-T0}), for finite-sample induced cut-off $k=k(T_0)$ when the generating model is trained on $T_0$ amount of data,
     $       E_{test} \asymp T^{-c} + T_0^{-c''}, 
     $
    where the exponents $c,c',c''$ only depend on the tail behavior of the true distribution. 
This result is illustrated in Figure \ref{fig:grokk}.

    For AI-''tail-narrowing", when data remains heavy-tailed, with a smaller exponent $\beta' \in (1,\beta)$, the downstream Hutter LLM  will scale as (Corollary \ref{cor:narrowtail})
    \begin{eqnarray}
    E_{test} \asymp T^{-(\beta-1)/\beta'}.
    \end{eqnarray}

\textbf{\textit{(2) A Triplet Scaling Law for Memory-Limited Models.}}
We consider a simple associative memory model studied in \cite{cabannes2023scaling}, and establish (Theorem \ref{thm:triplet}) a new scaling law of the form
    \begin{eqnarray}
    E_{test} \asymp T^{-c} + d^{-c_q} + k^{-c'},
    \label{eq:plateau3}
    \end{eqnarray}
where $d$ is the embedding dimension, and serves a s proxy for model capacity; the exponent $c_q$ depends both on $\beta$ and the particular algorithm $q$ used to update the memories in the model during training. 

\textbf{\textit{(3) Model Collapse over Multiple Generations.}} For $n$-fold recursion of AI data-generation \eqref{eq:gptchain}, where each generation of the model consumes data produced by the previous generation, we establish a universality principle of the form
\begin{eqnarray}
E_{test} = E_{test}^{clean} + \textcolor{red}{n \times \text{ new scaling terms}},
\end{eqnarray}
where $E_{test}^{clean}$ is the usual test error of the model trained on clean data (not AI-generated).
This means that in Equations \eqref{eq:plateau1} 
and \eqref{eq:plateau3} for example, the $k^{-c'}$ is replaced by $nk^{-c'}$. One possible interpretation of this multiplicative degradation is that, over time (i.e as the number of generations becomes large), the effect of large language models (like ChatGPT) in the wild will be a pollution of the web to the extend that learning will be impossible. 
We note that the multiplicative
degradation in scaling with the number of generations $n$ is analogous to what has been shown in \cite{dohmatob2024model} in the context of kernel ridge regression.
%

\textbf{\textit{(4)  Mitigation Strategies.}}
In Theorem \ref{thm:grokk} we show that mixing AI-generated data with even a small amount of clean data mitigates model collapse by introducing a grokking phenomenon. The length of the plateau is of order $k^\beta / \pi$, where $\pi$ is the proportion of training data which is from the true distribution (i.e clean data). When $\pi = 0$ (i.e only AI-generated data available), this plateau goes on forever (as in \eqref{eq:plateau1} and \eqref{eq:plateau3}). When $\pi > 0$, however small, the plateau finally halts, and the error continues to decrease à la $T^{-c}$. This grokking phenomenon holds in the setting of {\em deterministic} ground truth labels (like in the models of \citet{hutter2021learning,cabannes2023scaling}). For transformer models, such deterministic settings are found for instance in arithmetic tasks, and we demonstrate it empirically in our GCD transformer experiments. The grokking effect becomes attenuated in probabilistic settings, where it can lead to an S-shaped learning curve (see Figure \ref{fig:s-shape}). We also identify regimes where adding AI data can be beneficial and discuss ways to curate ''tail" data to mitigate AI-data effects.


\textbf{Related Work.} Theoretically, scaling laws have been derived in various settings: for non-parametric models \cite{SchmidtHieber2017scaling,suzuki2018adaptivity,BordelonCP20spectrum}, in the kernel regime under the Gaussian design \cite{Spigler_2020,cui2021generalization,Cui_2022,Cui_2023,maloney2022solvable}, or in memorization-like settings with discrete data \cite{hutter2021learning,debowski2023simplistic,michaud2023the}. Taking finite model capacity and optimization into account, \citet{cabannes2023scaling} recently proved scaling laws in constraint-capacity associative memories, and our Triplet Scaling Law builds on this work.

Less than a handful of works begin to provide theoretical explanations for the behavior of models in the ''synthetic data age".
\cite{shumailov2023curse} attribute model collapse to two mechanisms: a finite sampling bias cutting off low-probability ''tails", thus leading to more and more peaked distributions and function approximation errors; they theoretically analyze the (single) Gaussian case and provide empirical evidence for VAEs, Gaussian mixtures and the OPT language model (125M parameters). In the context of vision models, \citet{alemohammad2023selfconsuming} analyze {\em ''self-consuming loops"} by introducing a sampling bias that narrows the variance of the data at each generation, and, in addition to empirical demonstration on GANs and denoising diffusion probabilistic models, provide theoretical analysis for the Gaussian model.
Finally, let us mention the study of \citet{bertrand2023stability} which sheds light on the critical role of data composition in the stability and effectiveness in generative models, applicable to VAEs \cite{kingmaauto}, diffusion models and normalizing flows.
They explore scenarios involving a mix of clean data, representative of the true distribution, and synthesized data from previous iterations of the generator. Their analysis reveals that if the data mix consists exclusively of synthesized data, the generative process is likely to degenerate over time ({\em ''clueless generator"}). Using fixed-point analysis across iterations, they find that when the proportion of clean data in the mix is sufficiently high, the generator, under certain technical conditions, retains the capability to learn. A recent paper \cite{fan2023scaling} empirically observe deteriorated scaling laws when training on synthetic data for text-to-image models. \footnote{A more detailed description of related and prior work can be found in Appendix \ref{app:refs}} 



To our knowledge, our work is the first to theoretically and empirically analyze model collapse in the context of scaling laws and emergent abilities to provide a rich new landscape of AI-data induced phenomena. 


\vspace{-3pt}
\section{A Deterministic Infinite Memory Model}\label{sec:hutter}

Here, we present the core of our theory for the simplest case of (i) {\em infinite memory} and (ii) {\em a deterministic ground truth} labeling function $i \mapsto y_i$, studied by \citet{hutter2021learning} (the {\em ``Hutter LLM"}). Both restrictions will be lifted in later sections, where we also analyze an {\em probabilistic autoregressive} version (Section \ref{sec:bigram}) and {\em limited memory} models (Section \ref{sec:triplet}). Token $i$ is drawn according to the Zipf law in Equation~\eqref{eq:zipf}, which e.g. models distribution of various metrics in language. Another interpretation of the appearance of a power-law is offered by the ``Quantization Hypothesis" paper of \citet{michaud2023the}: one may think of each $i$ as some discrete skill, needed to solve a problem for example; thus, the skills occur at different rates $p_i$. The shape parameter $\beta > 1$ controls the length of the tail of this distribution: bigger values of $\beta$ correspond to longer tails. 

\subsection{What Causes Model Collapse ?}
\paragraph{Tail Cutting.} As mentioned, deliberate choices in the AI generation algorithm (like top-$p^{inf}$ or top-$k^{inf}$ next token prediction) immediately lead to a chopped tail at $k$. When viewed as skills, we can say that only the $k$th most frequent outcomes ("skills") are considered. But even when no tails are cut deliberately, the finite size $T_0$ of the training set (sampling bias)  induces an effective tail-cutting. This can be seen as follows:  Sample an iid dataset of size $T_0$, and estimate the histogram $p_{\text{AI}}$; this new distribution plays the role of an AI data-generator. An integer $i$  appears in the support of $p_{\text{AI}}$ a number of times which is $T_0p_i$ on average. Roughly speaking\footnote{This can be made rigorous via standard concentration arguments.}, this means that the support of $p_{\text{AI}}$ is $\{i \mid p_i \ge C / T_0\} = \{i \mid i \le k\}$, where 
\begin{equation}\label{eq:cutsample}
k = k(T_0) \asymp T_0^{1/\beta}.
\end{equation}
Therefore, the transformation $p \to p_{\text{AI}}$ amounts to chopping off the tail of $p$ at rank $k$, where $k$ is as given above. 

{\bf Tail Narrowing.} Figure \ref{fig:wiki-raw-test-evalgen} (for Llama2) shows that in addition to tail cutting, tail narrowing effects happen during AI-generation. One mechanism for this is lowered temperature during next-token prediction. Assume a softmax distribution on the logits $z_i$ for the $i$th token: $p_i=e^{z_i}/\sum_je^{z_j}$. Define $q_i^\tau=e^{z_i/\tau}/\sum_je^{z_j/\tau}$ for general temperature $\tau$. Then
$p_i \asymp i^{-\beta}$ morphs into $q_i^\tau \asymp i^{-\beta/\tau}$ (to first order). We see that temperature scaling directly causes narrowing of tail for $\tau>1$. Other mechanisms can come to play: for instance, for autoregressive models with perplexity, token-wise tail cutting can result in tail narrowing for sequence-perplexity (see Figure \ref{fig:cutoff=narrowing} and discussion in Appendix \ref{app:tail}).

\subsection{A New Scaling Law in the Hutter LLM}
For a deterministic ground-truth labelling function $i \mapsto j_i$, consider a downstream Hutter ``LLM" \cite{hutter2021learning}
\begin{eqnarray}
\label{eq:hutter}
    \widehat f(i) := \begin{cases}
        j_i,&\mbox{ if }(i,j_i) \in \mathcal D_T,\\
        \perp,&\mbox{ otherwise,}
    \end{cases}
\end{eqnarray}
constructed on a sample $\mathcal D_T := \{(i_t,j_{i_t}) \mid t \in [T]\}$ of size $T$ from unmitigated Zipf distribution $p$ \eqref{eq:zipf}, the test error obeys the following scaling law \citet{hutter2021learning}
\begin{eqnarray}
    E_{test} \asymp T^{-(1-1/\beta)}. 
    \label{eq:huttertest}
    \end{eqnarray}
Now, let $q$ be a k-tail-cutting version of $p$, i.e $q_i \propto p_i$ if $i \le k$ and $q_i=0$ otherwise. When constructed (``trained") on $\mathcal D_T$ of size $T$,  now from $q$, the test error (w.r.t to the true data distribution $p$) of this model is
\begin{eqnarray}
\begin{split}
E_{test} & := \mathbb P_{i \sim p}(\widehat f(i) \ne j_i)
= \sum_{i \ge 1} p_i \mathbb P(\widehat f(i) \ne j_i).
\end{split}
\end{eqnarray}
That is, we train on data from the AI distribution $q$ and test on original distribution $p$. We prove the following scaling law for tail cutting (all proofs are relegated to Appendix \ref{app:hutter}): 

\begin{theorem}
 Consider long-tail real-world data with exponent $\beta>1$, and let the cutoff for AI-generated data be $k$. Then, for large $k$ and $T$ samples from the AI, the test error of the downstream "LLM" scales like so $E_{test} \asymp T^{-(\beta-1)/\beta} + \textcolor{red}{k^{-(\beta-1)}} \asymp \min(T,\textcolor{red}{k^\beta})^{-(\beta-1)/\beta}$. 
 \vspace{-0.5cm}
 \label{thm:simple}
 \end{theorem}
Thus, as soon as $T \gtrsim k^\beta$, the AI-generated sample size $T$ ceases to be a "scalable" resource: collecting more AI-generated samples will not improve the performance of the downstream model, i.e performance plateaus and we lose scaling.
The result is illustrated empirically in  Figure \ref{fig:grokk}, left and Figure \ref{fig:simple} (Appendix \ref{app:figures}).


When we assume that the AI-generator itself was trained on $T_0$ samples, we get a similar loss of scaling stemming from the tail cutting from finite sampling bias (Equation \eqref{eq:cutsample}):
\begin{corollary}[``Finite Initial Sample Size"]
With $c=1-1/\beta$, it holds that
\begin{eqnarray}
\label{eq:basic}
    E_{test} 
    \asymp T^{-c} + T_0^{-c}.
\end{eqnarray}
\vspace{-0.3cm}
\label{cor:from-k-to-T0}
\end{corollary}
\vspace{-.5cm}
These theoretical are empirically confirmed in the Figure \ref{fig:chopping}. 
In the case of tail narrowing, the scaling behavior changes; instead of a plateau, we obtain a slower decay rate:

\begin{corollary}[``Tail Narrowing"]
    In the setting of Theorem \ref{thm:simple}, consider AI-generated data to also be long-tail data, albeit with smaller exponent $\beta' \in (1,\beta)$. Then, the downstream Hutter LLM trained on AI-generated data will scale as $E_{test} \asymp T^{-(\beta-1)/\beta'}$.
    \label{cor:narrowtail}
\end{corollary}


\subsection{Collapse Over Multiple Generations of AI Data}
\label{subsec:Cn}
We now examine the cumulative impact of prior loss of scaling across multiple generations. Consider $n$-fold recursive AI data-generation, i.e
\begin{eqnarray}
    p \to p_{\text{AI}(1)} \to p_{\text{AI}(2)}  \to \ldots \to p_{\text{AI}(n)}.
    \label{eq:gptchain}
\end{eqnarray}

Each arrow corresponds to drawing a sample of size $T_0$. If we iterate $n$ times the argument leading to \eqref{eq:basic}, we get the following scaling for the test error $E_{test}^{(n)}=E_{test}^{(n)}(T)$ for learning on $T$ samples from the $n$th generation and testing on the true data distribution, 
\begin{eqnarray}
\begin{split}
    E^{(n)}_{test} &\asymp T^{-c} + \underbrace{T_0^{-c} + \ldots + T_0^{-c}}_{n \text{ times}}\\
    &= T^{-c} + n T_0^{-c} = T^{-c}\left(n(T/T_0)^c + 1\right),
\end{split}
    \label{eq:nfold}
\end{eqnarray}
where $c:=1-1/\beta$. Only in the context of model collapse across multiple generations, $T$ is the size of the AI data in the final step and $T_0$ is the data a model trained in all preceding steps. We deduce the following result.

\begin{theorem}[Informal]
Model collapse (as spoken of in the literature) occurs iff $n \gg (T_0/T)^c$.

For example, if $T_0 \gg T$  (e.g $T_0 \ge C T\log T$) and $n$ is constant (e.g $n=25$), then model collapse will not occur if we learn on the $n$th generation of AI data. On the other hand, if $T_0 \lesssim T$, then model collapse will eventually occur.
\label{thm:n_law}
\end{theorem}
In particular, taking $T_0 \asymp T$, we get
\begin{eqnarray}\label{eq:n-law}
    E^{(n)}_{test} \asymp C_n T^{-c} \asymp \textcolor{red}{n} T^{-c}.
\end{eqnarray}
Note how the loss scales linearly with the number of generations. Figure \ref{fig:grokk}, middle, illustrates how an increased number of generations moves the loss scaling curve progressively to the right. This leads to eventual model collapse.



\vspace{-4pt}
\section{Mitigating Model Collapse via Data Mixing}\label{sec:grok}


Here we explore the possibility of alleviating model collapse via the acquisition of even a tiny amount of data from the true data distribution, to complement AI polluted data. We study two phenomena: (1) In the case of mixing $\pi$-fraction of the original data with a $(1-\pi)$ fraction of AI-generated data we exhibit a startling ``grokking" phenomenon where test loss plateaus with increasing training data to finally decrease again according to the scaling law of the original model, and (2) in the scenario where we would like to compensate for missing ``tail", we acquire some data from the tail of the original distribution to show that this needs to be done with caution: getting data from ``too deep" in the tail is worthless while data closer to the precise ``missing" tail can be beneficial. All proofs can be found in Appendix \ref{app:hutter}.

\vspace{-2pt}
\subsection{Acquiring Missing Tail}

To counter the effect of tail cutting and the resulting plateau in scaling, we might resort to adding curated data that would emphasize the tail. The following Theorem \ref{thm:annealed} studies this effect; it shows, in particular that if we ``overshoot" and only curate tail that is too deep, our efforts will be worthless. Rather, there is a fine line around the chopped tail $k$ (within a factor of $(1+o(1))$ of $k$), where we need to place our data curation efforts to get the scaling back.

 Suppose we ``buy" a chunk of the tail of the real data distribution corresponding to $i=N,N+1,\ldots$; let the distribution be $\pi$ (thus, supported on $\{N,N+1,\ldots\}$).
 Now, let $k$, $N$, and $T$ tend to infinity such that $N/k \to C$, with  $C \in [1,\infty]$.
 We have the following sharp phase-transition.
\begin{theorem}
(A) If $C=1$, e.g if $N=k + \sqrt k$, then $E_{test} \asymp T^{-c}$. That is, we perfectly anneal the tail-chopping effect of AI-generated data.

(B) If $C > 1$, then
$E_{test} \asymp T^{-c} + k^{-\alpha}$ (which recovers the result of Theorem \ref{thm:simple}), and so ``buying" the $N$th tail of the real data distribution is worthless.
\label{thm:annealed}
\end{theorem}

\vspace{-2pt}
\subsection{A Grokking Phenomenon}


Here we show how even small amounts of original data can mitigate the above ``scaling law collapse" by introducing a grokking phenomenon where test error plateaus and eventually continues to decline.


\begin{theorem}[Grokking with Tail Cutting]
Consider a sample of size $T$ of which a proportion $\pi$ comes from the true distribution $p$ and the remainder comes from a version $p'$ of $p$ with its tail chopped off at rank $k$. We have the following scaling laws for the Hutter LLM define din \eqref{eq:hutter}.

(A) \textbf{Early-Stage Dynamics.} For $T \ll k^\beta/\pi$, it holds that
\begin{eqnarray}
    E_{test} \asymp T^{-(1-1/\beta)} + k^{-(\beta-1)}.
\end{eqnarray}
Thus,  during this stage, the money spent on acquiring some clean data is not amortized!

(B) \textbf{Later-Stage Dynamics.} As soon as $ T \ge Ck^\beta/\pi $ (where $C$ is an absolute constant), it holds that
    \begin{eqnarray}
        E_{test} \asymp (\pi T)^{-(1-1/\beta)}.
    \end{eqnarray}
Thus, during this stage, we recover the unpolluted  sample-size law scaling $T^{-(1-1/\beta)}$, up to within a multiplicative constant $\pi^{-(1-1/\beta)}$ (which can be seen as an increase in the price of data). For fixed $T$ and tunable $\pi$, this error rate scales like $\pi^{-(1-1/\beta)}$, which is yet another scaling law.
\label{thm:grokk}
\end{theorem}

Effectively, the above theorem predicts that for any fixed $\pi \in (0,1)$ --no matter how small-- the test error grokks w.r.t sample size $T$. The result is empirically confirmed in Figure~\ref{fig:grokk}, right (see Figure \ref{fig:grokk-old} for another illustration).

 We experimentally confirm this new phenomenon for transformer models trained to calculate the GCD (see Appendix \ref{app:gcd}), which indicates its applicability for a wider class of LLMs with underlying deterministic ground truth, like for arithmetic tasks.

In Appendix \ref{app:groknarrow} we state and prove a similar theorem in the case of {\em tail narrowing} of synthetic data.

 \paragraph{Benefits of Mixing with AI Data.} The above machinery allows us to analyze a particular regime where AI-data can help improve performance. 

Taking $T=T_{real}+T_{AI}$ and $\pi=T_{real}/T$, we have the following important corollary of Theorem \ref{thm:grokk}. 
\begin{corollary}
For $T_{real} \ll k^\beta$, it holds that
    $E_{test} \asymp (T_{real} + T_{AI})^{-(1-1/\beta)} + k^{-(\beta-1)}.$
\label{cor:bump}
\end{corollary}

\begin{figure}[tb]
\vspace{-6pt}
    \centering
    \includegraphics[width=0.7\linewidth]{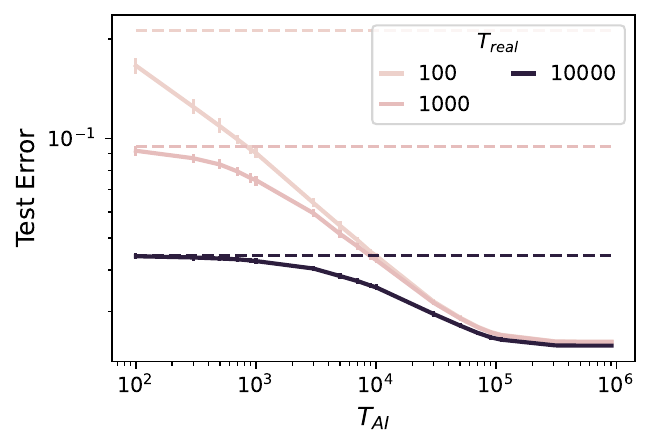}
    \vspace{-0.5cm}
    \caption{Mixing $T_{real}$ real data with $T_{AI}$ AI data. The dotted lines depict test errors of real data alone. $k=1,000, \beta=3/2$.}
    \label{fig:benefits of ai}
    \vspace{-0.4cm}
\end{figure}

Figure \ref{fig:benefits of ai} illustrates how AI data can boost performance, up to a certain point, when its benefits plateau.
This result might contribute to our understanding of why, sometimes, adding AI-generated data might lead to better models, especially when generated by a stronger model (e.g.~\citet{he2023is,shipard2023diversity,bansal2023leaving,10208358}). See Appendix \ref{app:refs} for more references. 





\section{A Tailed Bigram Model}\label{sec:bigram}



We will now proceed to a more complex model, bringing us closer to capturing the {\em probabilistic} and {\em autoregressive} nature of LLMs (next token prediction). 
In this Section we will define the data generating process, define the new model (Hutter++), and establish that the original scaling law (with clean data) still holds. We then proceed to show similar loss of scaling for AI-data.

A first fundamental step is to consider {\em probabilistic} ground truth labels to replace the deterministic Hutter prediction $i \mapsto y_i$ with a probability distribution $p(j|i)$ on $\mathbb N_*$
with  power law decay (as in Equation \eqref{eq:zipf}).
 To account for the fact that the most frequent next token $j$ depends on the preceding token $i$ we model 
\begin{eqnarray}
\label{eq:tailoutput}
p(j \mid i) \propto \pi_i(j)^{-\beta},
\end{eqnarray}
(instead of $j^{-\beta}$),  where $\pi_i$ is a permutation associated to every $i$ providing the order of outputs. 
To summarize, we think of the data as pairs $(i,j)$, where the distribution of $i$ is governed by some $p(i)$ as in the deterministic Hutter setting, and $p(j|i)$ is given by Equation~\eqref{eq:tailoutput}.

This setting can be made {\em autoregressive} by generating sequences step by step, using the preceding output as the next input. We can think of each successive pair of tokens as of the pairs $(i,j)$ above, with the only difference that the marginal distribution $p(i)$ changes. We thus will make no assumptions on $p(i)$ in what follows (except for a mild technical condition). Proofs can be found in Appendix \ref{app:bigram}.

\paragraph{Remark.} The Hutter model abstracts Large Language Models (LLMs) into encapsulated knowledge or skills \citet{michaud2023the}, denoted as $i$. This variable, $i$, can be interpreted as a composite description comprising multiple components, where the Hutter model highlights the tendency of AI-generated data to exhibit a less pronounced heavy-tailed distribution. In scenarios where AI-generated data encompasses both accurate and erroneous information—such as in predictions of the greatest common divisor—a bias towards incorrect responses is observed. The Hutter++ model extends this framework by incorporating this specific bias. Specifically, we now curate the model to generate AI data with the consistent probability distribution $p(i))$. Errors are represented within the conditional distribution $p(j|i)$, which, in turn, affects the model's performance in later training stages. 


\subsection{The Hutter++ Algorithm}
We now present an extension of the Hutter model \eqref{eq:hutter} which is adapted to bigrams. Let $n_T(i) = \sum_{t=1}^T 1[i_t = i]$ be the number times the context $i_t$ appears in the dataset $\mathcal D_T$ and 
$n_T(i,j) = \sum_{t=1}^T 1[(i_t,j_t) = (i,j)]$ be the number of times the pair $(i,j)$ appears in the dataset. Note that $n_T(i) \sim Bin(T,p_i)$. As soon as $n_T(i) \ge 1$, define
$$
q_T(j \mid i) := n_T(i,j) / n_T(i).
$$
This is an empirical version of $p( \cdot \mid i)$ based on an iid sample of size $n_T(i)$.  For a theoretical analysis, we shall consider the following test error metric based on total-variation (TV)
\begin{eqnarray}
\label{eq:TVloss}
E_{test} := \sum_i p_i\, \mathbb E\, [TV(q_T(\cdot \mid i), p(\cdot \mid i))],
\end{eqnarray}
where $TV(a, b) := \sum_j |a_j-b_j|$ is the total-variation distance and the expectation is over the randomness in $q_T$. 
An asset  here is that \cite{Berend2012OnTC} can be used to control the quantities $\mathbb E\, [TV(q_T(\cdot \mid i), p(\cdot \mid i))]$. 
Note that TV is upper-bounded by the square-root of KL-divergence, thanks to \emph{Pinker's inequality.} This gives indication that our results could also apply in the setting of autoregressive models with perplexity loss, like LLMs.


\subsection{A Scaling Law for Hutter++}
Consider a case of non-deterministic outputs as in Equation \ref{eq:tailoutput}, where $\pi_1,\pi_2,\ldots$ are functions
from $\mathbb N_*$ to $\mathbb N_*$.
\begin{theorem}
\label{thm:bigram1}
    Suppose $\beta \in (1,\infty)\setminus\{2\}$ and set $c:=\min(1-1/\beta,1/2)$. If $\sum_i p_i^{1-c} < \infty$, then $E_{test} \lesssim T^{-c}.$
    Moreover, if $\beta \in (1,2)$ and the mappings $\pi_1,\pi_2,\ldots$ are permutations, then $E_{test} \asymp T^{-c}$.
    \vspace{-0.3cm}
\end{theorem}
Thus, the proposed Hutter++ algorithm induces exactly the same scaling law as the classical setup \cite{hutter2021learning} !

\subsection{Model Collapse in Probabilistic Setting}
We now return to our main problem, understanding model collapse in the probabilistic setting and  consider the Hutter++ presented above. Thus, suppose the learner only has access to at most a dataset of size $T$ containing the $k$th head of the conditional distribution $p(\cdot \mid i)$. That is, sampled from: $i \sim p$,  $j \sim p(j \mid i)1[j \le k]$ (normalized appropriately), where $p(\cdot \mid i)$ is as in Equation~\eqref{eq:tailoutput}.
\begin{theorem}
\label{thm:bigram3}
   (A) If $\beta \in (1,\infty)\setminus \{2\}$ and $\sum_i p_i^{1-c} < \infty$ where $c:=\min(1-1/\beta,1/2)$ as before, then $E_{test} \lesssim T^{-c} + k^{-\beta c}.$
 
(B) Furthermore, if the mappings $\pi_1,\pi_2,\ldots$ are permutations and $\sum_i p_i^{1-c} < \infty$, then $E_{test} \asymp T^{-c} + k^{-\beta c}$.

\end{theorem}

{\bf Autoregressive Bigrams.} Similarly these results hold for autoregressive bigram model, where $p(i_1,i_2,\ldots,i_L) = p(i_1)\prod_{\ell=1}^{L-1} p(i_{\ell+1} \mid i_\ell)$, and each $p(j \mid i)$ is as in \eqref{eq:tailoutput}.
The result is empirically confirmed in Figure \ref{fig:bigram3} in Appendix \ref{app:figures}.

{\bf Multiple Generations.} The mechanics of the proof of Theorem \ref{thm:n_law} apply in this setting. 
See Figure \ref{fig:bigram_nocutoff} in Appendix \ref{app:figures} illustrating that Equation \eqref{eq:n-law} keeps holding for probabilistic data distributions.

{\bf Grokking for Mixtures.} Technically speaking, this grokking phenomenon only holds for models with deterministic ground truth labels, like the Hutter LLM and the limited capacity associative memory model. For the {\em probabilistic} setting of bigrams (or text LLMs) the theorem cannot hold in its pure form, because if we train on a mixture of two distributions (clean and synthetic) but test only on the clean distribution, the distance between these two distributions will always be a lower bound on the test error. However, we can see that remnants of a ``smoothed" grokking-law persist in the form of an S-shaped scaling (see Figure \ref{fig:s-shape}).



\vspace{-3pt}
\section{Capacity-Limited Memory Models: A Triplet Scaling Law}\label{sec:triplet}

We now turn to a finite-memory extension of the Hutter LLM, which allows to model {\em capacity}. 
We look into a simple associative memory model \cite{cabannes2023scaling}:
\begin{eqnarray}
\label{eq:eva}
\begin{split}
    f_T(i) &:= \arg\max_y H_T(i,y),\text{ where}\\
    H_T(i,y) &:= e_i^\top M_T u_y,\\
    M_T &:= \sum_i q_T(i) e_i u_{f_\star(i)}^\top \in \mathbb R^{d \times d}.
    \end{split}
\vspace{-0.3cm}
\end{eqnarray}
This is a transformer-like finite-memory extension of 
 the infinite-memory model in \cite{hutter2021learning}. The integer $d \ge 1$ then plays the role of the "capacity" of the resulting model. Here, $f_\star:[N] \to [m]$ is an unknown function, for example, reduction modulo $m$, i.e $f_\star(i) := ((i-1)\text{ mod } m) + 1$; $q_T=q(\mathcal D_T)$ is probability distribution on $[N]$ which encodes an arbitrary learner, estimated using and iid sample $\mathcal D_t = \{(i_t,y_t) \mid t \in [T]\}$ of size $T$ collected from a probability distribution on $[N] \times [m]$, of the form
\begin{eqnarray}
    i \sim p = \mathrm{Zipf}(\beta),\quad y = f_\star(i).
\end{eqnarray}
The embedding vectors $e_1,e_2,\ldots e_N$ and $u_1,u_2,\ldots,u_m$ are a system of unit-vectors in $\mathbb R^d$, constructed so that the matrix $\mathbb R^{d \times d}$ remembers the input/output pairs $(i,j)$ it has seen, i.e $e_i^\top M u_{f_\star(i)} \approx q_T(i)$ if $(i,f_\star(i)) \in \mathcal D_T$. The weights $q_T(i)$ ensure that different memories are memorized faster than others.

\citet{cabannes2023scaling} proposed that iid random embeddings from the uniform distribution on the unit-sphere in $\mathbb R^d$ be used. In this setting, for different choices of $q$, the following general scaling law was established
\begin{eqnarray}
    E_{test} \asymp T^{-(1-1/\beta)} + d^{-c_q},
\end{eqnarray}
where the exponent $c_q \in (0,\infty)$ depends on $\beta$ and the algorithm $q$. For example, when $q$ encodes the counting measure $q_T(i) := n_T(i)/T$ (reminiscent of SGD), it was shown that $c_q = (1-1/\beta)/2 \in (0,1/2)$. Another algorithm  $q_T(i) := 1[n_T(i) \ge 1]/\sum_\ell 1[n_T(\ell) \ge 1]$ (remniscent of ADAM ) was proposed which attains a optimal error rate (over all algorithms based on random embeddings) with $c_q = \beta-1$.

In the context of model collapse which is the main focus of this manuscript, we have the following.
\begin{theorem}[Triplet Scaling Law]
For all the algorithms $q$ considered in \cite{cabannes2023scaling}, one has the following triplet scaling law w.r.t sample size $T$, embedding dimension $d$, and frequency cutoff $k$,
    \begin{eqnarray}
        E_{test} \asymp T^{-(1-1/\beta)} + d^{-c_q} + \textcolor{red}{k^{-(\beta-1)}}.
    \end{eqnarray}
    \vspace{-0.3cm}
    \label{thm:triplet}
\vspace{-0.7cm}
\end{theorem}
This result is empirically confirmed in Figure \ref{fig:tripletlaw}  and proved in Appendix \ref{app:triplet}. It gives rise to similarly tapered-off scaling curves for synthetic data, as in the simpler models.
The proofs for loss of scaling across generations in Section \ref{sec:hutter} and grokking phenomena in Section \ref{sec:grok}, carry over to this model as well, demonstrating their universality.

\vspace{-4pt}
\section{Experiments}\label{sec:experiments}

In this section we present our experimental results to demonstrate evidence of various predictions we have made theoretically. We showcase four scenarios of increasing level of complexity: an empirical Hutter++ model, autoregressive bigram models with {\em perplexity loss}, an arithmetic transformer to predict the GCD of two integers \cite{charton2023transformers} and a large-scale LLM, Llama2-7B \cite{touvron2023llama}, trained on a large data corpus ({\tt Wikidata-103}).


In our theoretical analysis, motivated by empirical observations (see Figure \ref{fig:wiki-raw-test-evalgen}) or by the effect of finite data sampling bias on heavy-tailed data, we have assumed that generated data follows patterns of either a tail cutoff or tail narrowing. 
In our subsequent experiments, we depart from theoretical assumptions on tail-cutting/narrowing to allow the widely deployed top-$p^{inf}$ selection or temperature scaling mechanisms to give possibly intermingled effects on the generated data distribution.

\begin{figure}[htb]
    \centering
    \vspace{-6pt}
     \includegraphics[width=0.7\linewidth]{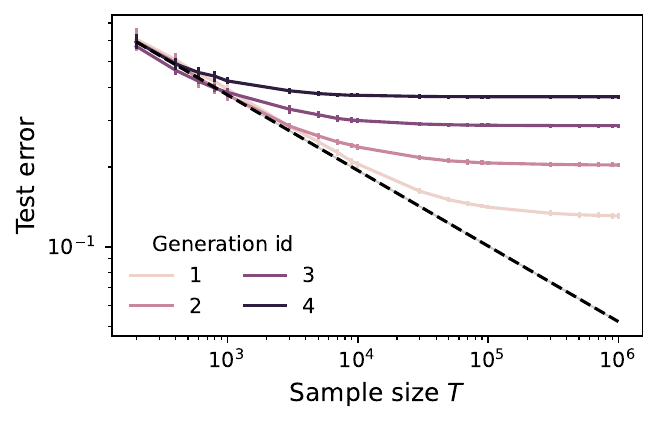}
     \vspace{-0.5cm}
    \caption{{\bf Hutter++ on Bigram with limited data and top-$p^{inf}$.} The initial model is trained on $T_0=100,000$ samples. It generates $T$ samples for Gen 1. Starting from Gen 2 models are trained on data generated by the most powerful model from the previous generation. Top-$p^{inf}=$0.95 cutting and $\beta=3/2$.}
    \label{fig:top-p-95}
    \vspace{-0.8cm}
\end{figure}

{\bf Empirical Hutter++ Model.} In Figure \ref{fig:top-p-95}, we use an initial model that is trained on $T_0=100,000$ samples from the original distribution. For the Gen 1 line, the data are all generated from this initial model. From Gen 2 onwards, models are iteratively trained on data produced by the most performant model of the preceding generation, effectively eliminating the possibility that model collapse results from inadequate sampling. For Gen 1, a notable degradation in data scalability is observed, alongside a rapid decline in model performance across generations. These observations not only validate our theoretical result but also reaffirm our assumptions. A similar pattern is evident with temperature scaling, as shown in Figure \ref{fig:temp90}.


{\bf Autoregressive Bigram Models with Perplexity Loss.}
We move one step further towards ``real" LLMs to investigate autoregressive bigram models. The dataset now comprises sequentially generated integers, adhering to Equation \eqref{eq:tailoutput}, with the model trained on all tokens. We use the averaged perplexity score of the test set as the test error metric. Our study encompasses a range of effects—such as top-$p^{inf}$ inference, temperature scaling, limited real data, and training on progressively larger AI datasets. Consistent with the findings in Section \ref{sec:bigram}, we observe the same patterns of scaling loss and progressive model collapse across generations. Relevant figures are provided in Appendix \ref{app:perplexity}.

{\bf Transformers Learning the GCD.}
Our first illustration of our theory ``in the wild" is for sequence-to-sequence transformer models for an arithmetic task: predicting the greatest common divisor (GCD) of two integers, encoded as sequences of digits in some base $B$, following \citet{charton2023transformers}. This setup is a perfect intermediate step between our toy models and large scale LLMs; it uses the transformer architecture and training algorithms on sizeable models, while the underlying data has a deterministic nature.
Over the course of training the model progressively learns new GCDs and with them also their products with already learned GCDs. We can thus view each such learned group, usually learned in ``bursts", as a new skill. 
For the purpose of this experiment, we use this model after $300M$ samples as the generator of AI-data. In Figure \ref{fig:experiments} 
we validate the predicted scaling law for a single generation and observe `un-learning' of skills when training exclusively with generated data, as well as a grokking effect when training on mixtures. See Appendix \ref{app:gcd} for details and more figures.

{\bf Experiments on LLMs.}
We finetune Llama2 with LoRA, generating synthetic AI data for the next finetuning iteration. Inspired by the setup in \citet{shumailov2023curse}, we use {\tt Wikidata-103}, partitioned into approximately $2.2$ million sequences of 128 tokens. AI data is generated through prompt completion, using the first 96 tokens from the original sequences as prompts. The model is trained only on the last 32 tokens to preclude information leakage. 
The evaluations are conducted exclusively on the same 32 tokens. We use top-$p^{inf}=$0.9 and temperature $\tau=$0.9 across all generations. The results, depicted in Figure \ref{fig:experiments} (left), illustrate a scaling law decay over several generations. The first generated dataset still contains useful but limited information; the utility of the second generation's data markedly diminishes. These phenomena corroborate the anticipated loss of scaling law and model collapse, further indicating that model collapse is even more pronounced here, 
highlighting the challenges in training next generation LLMs. More details and results in Appendix \ref{app:llama}.

\begin{figure}[htb]
    \centering
    \vspace{-6pt}
     \includegraphics[width=0.65\linewidth]{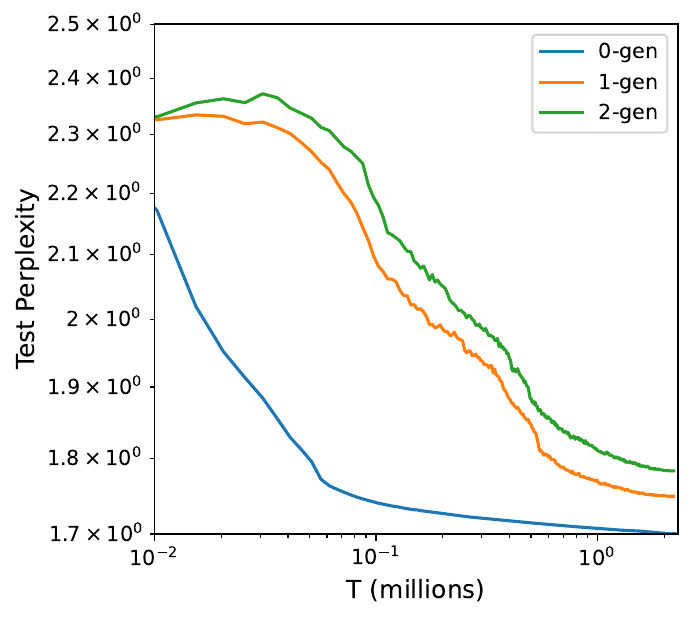}
     \vspace{-0.5cm}
    \caption{{\bf Mixing Llama Generated Data with Original Data} Based on Figure \ref{fig:experiments} left, we mix generated data with original data, with ratio 98 to 2. Note how the mixing curve validates our predicted curve of the grokking phenomenon as in Figure \ref{fig:grokk}}
    \label{fig:llama_mix_02}
    \vspace{-0.5cm}
\end{figure}

Moreover, we conduct experiments to investigate mixing a proportion of real data with AI-generated data. Figure \ref{fig:llama_mix_02} demonstrates the effect of blending a random 2\% of original data with AI data across all fine-tuning phases. It significantly mitigates model collapse, with the emergence of a grokking curve as predicted in Theorem \ref{thm:grokk}.

\section{Conclusion}

In the advent of the ``synthetic data age", our work signals the end of current neural scaling laws and opens the door to a puzzling array of new phenomena in a world where the training corpora are enriched with AI generated data. 
We demonstrate that scaling laws cease to persist; test error tapers off due to altered, less heavy tailed, data distributions. 
Yet, new opportunities arise from careful mixture and data curation, as we have shown, with interesting effects at the interplay of clean and synthesized data. We must recognize new learning plateaus and, for instance, adjust to changed learning curves from blending clean and synthetic data to unintended early stopping. 
A notable feature of our work is that our theory is {\em effective} - we observe the predicted phenomena for relevant large models in two different settings.


Our contributions call for a more responsible, or ``collapse-aware", proliferation of synthesized data. Scale is \textit{not} all you need: more work on watermarking for synthetic data is needed, to make it more distinguishable from original human-annotated data. ``Real" data will become an even more valuable resource in the future, as we are ushering in the  ``beyond scaling" era.


\section*{Acknowledgements}

YF and JK acknowledge support through NSF NRT training grant award 1922658. YF and PY would like to thank Di He for
discussions and suggestions.
This work was supported in part through the NYU IT High Performance Computing resources, services, and staff expertise.


\section*{Impact Statement}

This study contributes to the discourse on AI scaling laws by tackling the emergent challenge presented by synthetic data within training datasets. It holds particular relevance for fields that depend heavily on data accuracy, such as natural language processing and generative AI model development. We investigate the concept of "model collapse" as a cautionary tale, highlighting the potential risks to sustainable model performance in the face of an increasing reliance on synthetic data. Through a combination of theoretical analysis and empirical evidence, we advocate for the strategic integration of clean data to mitigate these risks and enhance the robustness of AI systems. Our findings carry significant implications for industries where decision-making is progressively data-driven, suggesting a need for a shift in data management strategies to emphasize the preservation of model integrity and longevity. Additionally, this research highlights a looming issue: the potential scarcity of clean data. It underscores the urgency of focusing on the cultivation and preservation of high-quality datasets for AI training, to safeguard the future of AI development.

\bibliography{main}
\bibliographystyle{icml2024}

\newpage

\appendix
\onecolumn

\section{Prior Work}\label{app:refs}



\paragraph{Model Collapse:} \citet{lebrun2021evaluating} first investigate training on AI-generated texts using transformers and LSTMs, revealing the distributional distortion inherent in these neural language models. The phenomenon of model collapse is first proposed by \citet{shumailov2023curse} and has recently appeared in the literature in the context of language and image generation. Several recent works demonstrate facets of this phenomenon {\em empirically} in various settings \cite{Hataya_2023_ICCV, martínez2023combining,martínez2023understanding,bohacek2023nepotistically,briesch2023large,guo2023curious,fan2023scaling}. Only few recent works also provide some accompanying theoretical analysis \cite{shumailov2023curse,alemohammad2023selfconsuming,bertrand2023stability} which we outline now.

\citet{shumailov2023curse} define model collapse and attribute it to two mechanisms: finite sampling when training a model (leading to cut off of low-probability data) and function approximation errors (the model is not sufficiently expressive to model the true distribution). They observe (and, for a single Gaussian, prove) that upon iteratively resampling finite ``training data" the generated distribution becomes more and more peaked. Other models studied empirically are mixtures of (two) Gaussians and VAEs on MNIST. To study language models, \citet{shumailov2023curse} iteratively fine tune Meta's OPT-125M model on {\tt wikidata2}. For generation of new text they use a 5-way beam search, which, by its nature, (approximatively) generates only low-perplexity data. 



 \citet{alemohammad2023selfconsuming}  conduct an empirical and analytical analysis on generative image models of what they term the ``self-consuming" or ``autophaguous" loop. They conclude that without enough fresh real data at each generation, future models necessarily will have their precision or recall decrease. They model the influence of each new AI-generation via a generic {\em sampling bias} $0 \leq \lambda \leq 1 $. In the case of image generation this refers to feature parameters at generation that favor quality over diversity (suitably quantified). More precisely, $\lambda=1$ corresponds to unbiased sampling and $\lambda=0$ corresponds to sampling from the modes of the generative distribution. $\lambda$  models biased sampling methods commonly used in generative modeling practice, such as truncation in BigGAN and StyleGAN or guidance in diffusion models. 
 In the case of Gaussian distributions, $\lambda$ is the shrinking factor of the variance of the next generation. Their empirical work studies GANs and denoising diffusion probabilistic models for image generation on FFHQ and MNIST and single Gaussians for both theoretical and empirical observations. As in \citep{shumailov2023curse} they observe (and prove for the case of a single Gaussian) that estimation error alone leads to vanishing variance with number of iterations.
 \citet{alemohammad2023selfconsuming} also empirically observe an initial boost in performance in a regime where modest amounts of synthetic data are mixed with the original data before larger amounts of synthetic data lead to ultimate degradation. This might mimick larger-scale results that demonstrate how synthetic data mixed with true data improves performance in some scenarios (see {\em Benefits of synthesized data} below). Indeed, in its simplest form, data augmentation (rotations, cropping etc.~), a widespread highly beneficial practice in ML training, can be viewed as the simplest form of data generation.

Let us mention the study of \citet{bertrand2023stability} in the context of image generation, which sheds light on the critical role of data composition in the stability and effectiveness in generative models. They explore scenarios involving a mix of clean data, representative of the true distribution, and synthesized data from previous iterations of the generator. Their analysis reveals that if the data mix consists exclusively of synthesized data, the generative process is likely to degenerate over time, leading to what they describe as a `clueless generator'. Thus, the generator collapses: it progressively loses its ability to capture the essence of the data distribution it was intended to model. Conversely, they found that when the proportion of clean data in the mix is sufficiently high, the generator, under certain technical conditions, retains the capability to learn and accurately reflect the true data distribution. This work sheds light on the critical role of data composition in the stability and effectiveness of generative models.

Several empirical studies confirm the deleterious effect of training on self-generated data: In the context of image generation, 
\citet{martínez2023combining,martínez2023understanding} report  degradation of models trained on AI-generated data. Specifically, they use a Denoising Diffusion Implicit Model and a few (relatively small) datasets (e.g.~Orchids, MNIST) to demonstrate visual degradation when training in successive generations of AI-generated data. 
\citet{Hataya_2023_ICCV} {\em ``conclude that generated images negatively affect downstream performance, while the significance depends on tasks and the amount of generated images"}, \citet{bohacek2023nepotistically} reports that the popular StableDiffusion model collapses when iteratively retrained on self-generated faces, even with as little as $3\%$ synthetic data mixed into the original training set. For text, \citet{briesch2023large} use {\em nanoGPT}\footnote{\tt https://github.com/karpathy/nanoGPT} on a curated 10K logical-expression dataset to demonstrate the iterative collapse of self-consuming loops - the model and dataset are sufficiently small to allow training from scratch. \citet{guo2023curious} observe a decline in linguistic diversity metrics across iteratively fine-tuned LLMs. 

\paragraph{Mitigation:} To our knowledge, rigorous theory (or even empirical demonstrations) on mitigation strategies against model collapse are yet to come, with one notable exception in \citep{bertrand2023stability} (see below). Several works discuss the need for {\em detection} of AI-generated images or text (to avoid retraining on them), for example motivating research into watermarking strategies. 
\citet{bertrand2023stability} analyze iterative retraining on a mixture of synthesized and original data under several technical assumptions and find that there are fixed points governing the stability of iterative retraining.

\paragraph{Benefits of Synthesized Data} There is a range of results showing benefits of AI-synthesized data in training better models, though mostly these results pertain to image data, specifically in the context of diffusion models \citep{azizi2023synthetic,he2023is,shipard2023diversity,bansal2023leaving,10208358}, though not only (see \citet{dai2023auggpt,xu2023baize,huang2022large,wang-etal-2023-self-instruct} for chat-related examples). One might argue that they either throw model-collapse caution to the winds or, possibly, settle in the protected corner where mild amounts of synthetic data (or larger amounts of ``mildly synthetic" data, like in the case of data augmentation) helps. In particular, often benefits of synthetic data are observed when the synthetic data is generated by a model trained for a different use case than the downstream task (like images synthesized from diffusion models helping classification models) or generated by a stronger model \cite{he2023is,shipard2023diversity,bansal2023leaving,10208358}.
However, other works critically analyze the purported benefit of generated data. \citet{burg2023image} find that while synthesized data from a diffusion model helps improving downstream tasks, such as classification, using the {\em pre-training data} of the diffusion model alone gives even stronger performance (which we can interpret as evidence of mild first-generation model collapse). All in all it is fair to say that the impact of data augmentation using generative models is still not fully understood.

\paragraph{Scaling Laws:}
Neural scaling laws have been ubiquitously observed in vision, language and speech. 
Early large scale empirical studies are performed in \cite{hestness2017deep,rosenfeld2020a}, demonstrating power law scaling across a range of learning scenarios. This is followed by  well-known large-scale studies from OpenAI
\cite{kaplan2020scaling} and DeepMind \cite{hoffmann2022trainingChinchilla}, which empirically demonstrate power-law scaling in LLMs across a wide set of scales. Essentially, this empirically establishes that 
$$L(N,D) \sim N_C\cdot N^{-\alpha_N}+D_C \cdot D^{-\alpha_D},$$ where $L$ is the per-token cross entropy loss (in nats), $N,D$ are the number of (non-embedding) parameters and data, respectively, and $N_C, D_C$ and $\alpha_N,\alpha_D$ are constants determined by the data distribution and the model specifications. 

This study was extended  to demonstrate many more power law relations in various scenarios (vision transformer, video modeling, multimodal models, and mathematical problem solving) \cite{henighan2021scaling}. In the machine translation (MT) setting, \citet{gordon-etal-2021-data} quantify scaling laws for standard benchmarks like BLEU and explain them via cross-entropy power-law scaling, thus positing a first universality of scaling laws across metrics. 
\citet{hernandez2021scaling} demonstrate similar empirical power-law scaling for transfer learning and \citet{aghajanyan2023scaling} provide a vast experimental body of evidence for scaling laws in mixed-modal language models.

However, a few results have nuanced the view of scaling as a panacea to improved loss. For instance, \citet{mckenzie2023inverse} present evidence for ''inverse sclaing" where flaws in the training objective or the data lead to U-shaped scaling. 

\paragraph{Theoretical Models for Scaling Laws:} 
From a theoretical angle, scaling laws have been shown analytically even before the emergence of large foundation models. For instance, \citet{Caponnetto2007OptimalRF} characterize the power-law generalization error of regularized least-squares kernel algorithms. The role of optimization can also be taken into account in this setting \cite{nitanda2021optimal}.
In the nonparametric literature, for example \citet{SchmidtHieber2017scaling} and \citet{suzuki2018adaptivity} derived the test error scaling of deep neural network in fitting certain target functions and \cite{BordelonCP20spectrum} analyze spectral dependence.

More recently, scaling laws have been shown for kernel models under the Gaussian design, e.g. in \cite{Spigler_2020,cui2021generalization,Cui_2022} for regression and \cite{Cui_2023} for classification. 
\citet{maloney2022solvable} study scaling laws for the random feature model in the context of regression. In the context of memorization for heavy-tailed data scaling laws have been shown in the infinite-memory setting \cite{hutter2021learning}, for ''quantized" skills \cite{michaud2023the} and for certain random data-generation processes \cite{debowski2023simplistic}. When taking model capacity and optimization into account, \citet{cabannes2023scaling} recently proved scaling laws in constraint-capacity associative memories.

To our knowledge, however, very few papers deal with the decay of scaling in the case of self-consuming loops.  A notable example is \cite{bartlett2020neurips} which studies iterated retraining in the context of self-(knowledge-)distillation in the kernel setting. However, this analysis is very distinct from our work, not only because it places itself in the kernel setting with Gaussian design, but also because it assumes the distillation setting, where the ''generation" stage is carefully optimized for the next stage training. In the case of synthesized data in the wild, this assumption can of course not be made. 

\paragraph{Emergence of ``Skills" and Scaling Laws:}

Scaling laws give us an insight on bang-for-the-buck style trade-off for model training. However, cross-entropy loss is not a goal in and of itself: we want to train models that are endowed with a larger and larger skill set as we scale them up. For instance,  \citet{gordon-etal-2021-data} provide intuition and empirics for the scaling of BLEU score for MT with cross-entropy loss as $$BLEU(L) \approx Ce^{-kL},$$ demonstrating ``emergence" of good BLEU performance with scale. This type of ``emergence" has been massively confirmed in \cite{wei2022emergent}, where a working definition of ``emerging" is ``not present in smaller models, but appears in larger models". In this sense, \citet{wei2022emergent} demonstrate empirically a large number of ``skills" appearing with scale, like Multi-Task NLU, Modular arithmetic, word unscrambling and transliteration.

A theoretical model, providing an underpinning of the necessity of scaling laws for the emergence of skill has recently been given by \cite{arora2023theory}. They analyse ``emergence" with the scaling laws as a departure point in a model that links cross-entropy loss in LLMs to basic skills to show that scaling laws enable the model to learn (and generalize) efficiently. 

Strengthening the tie between scaling laws and emergent skill, albeit in the {\em opposite} direction, \citet{michaud2023the} posit that skills that emerge in ``quanta" imply a scaling law of the loss. Related, \citet{chen2023skillit} assume a hierarchy of skills to derive data curation mechanisms to precipitate the emergence of skills, though they do not allude to scaling laws directly.

\section{Complimentary Figures for Sections \ref{sec:hutter}, \ref{sec:grok} and \ref{sec:bigram}}\label{app:figures}

\begin{figure}[h]
     \centering
     \begin{minipage}{0.48\textwidth}
     \includegraphics[width=\textwidth]{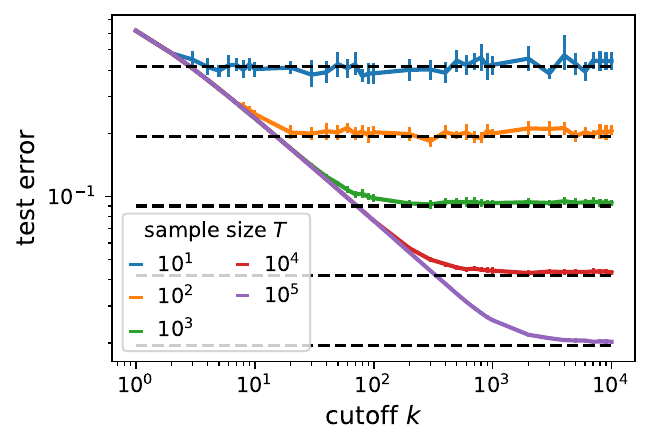}
    \vspace{-.5cm}
     \caption{\textbf{Scaling on Hutter LLM for Varying $T$.} Empirical confirmation of Theorem \ref{thm:simple}. Here, $\beta=3/2$ and error bars correspond to $10$ iid runs of sampling AI-generated data (i.e the distribution $q$).  Broken lines correspond to the Hutter rate $T^{-(\beta-1)/\beta}$, for varying $k$ and different values of $T$. Figure \ref{fig:grokk}, left, illustrates the same for varying $T$ and several settings of $k$. Note the perfect match with the theorem. }
     \label{fig:simple}
     \end{minipage}
     \hfill
     \begin{minipage}{0.48\textwidth}
        \includegraphics[width=\textwidth]{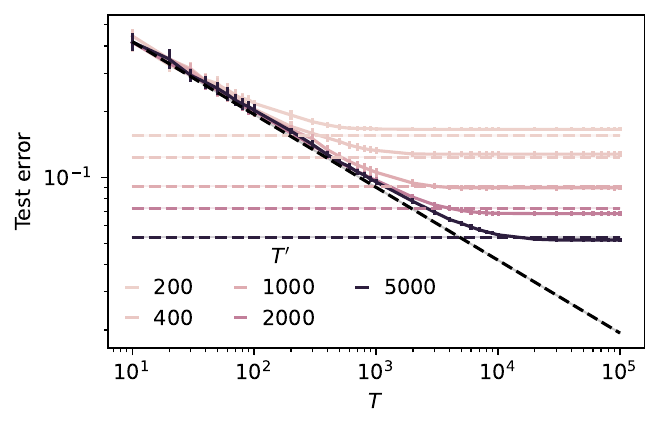}
    \vspace{-.5cm}
    \caption{\textbf{Scaling on Hutter LLM for Varying $k$.} A sample of size $T_0$ is used to approximate the true distribution $p$ via $p_{\text{AI}}$. Then, a Hutter-type model is learned on a sample of size $T$ from $p_{\text{AI}}$, and evaluated on the true data distribution $p$. Each horizontal line corresponds to the asymptote $k^{-\beta c} \asymp T_0^{-c}$, for different values of $T_0$. The diagonal line corresponds to $T^{-c}$.}
    \label{fig:chopping}
     \end{minipage}
 \end{figure}

\begin{figure*}[!h]
    \centering
    \includegraphics[width=1.\textwidth]{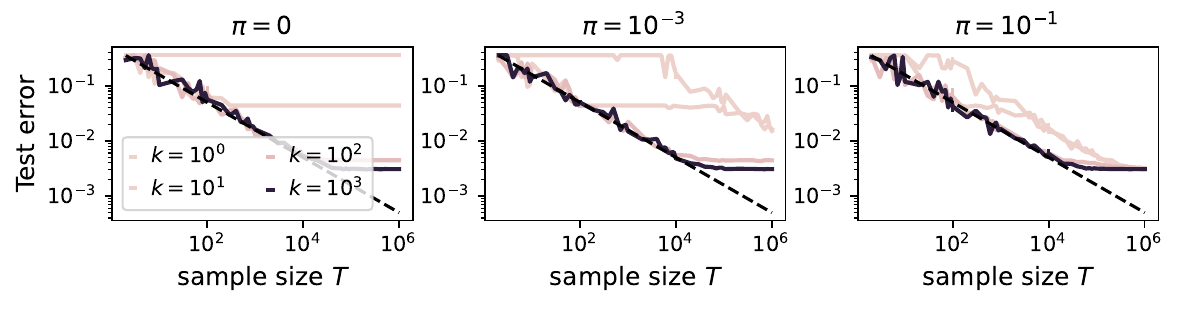}
    \vspace{-.75cm}
    \caption{Empirical Validation of Theorem \ref{thm:grokk}. The broken line corresponds to the $T^{-(1-1/\beta)}$ scaling law that would hold throughout in the absence of pollution. Notice the grokking behavior predicted by the theorem. For this experiment, the Zipf exponent of the true data distribution $p$ is $\beta=2$.}
    \label{fig:grokk-old}
\end{figure*}

\paragraph{Hutter LLM.} Figures \ref{fig:simple}, \ref{fig:chopping} and \ref{fig:grokk-old} further illustrate our theory for simple Hutter LLM.

\begin{figure}[!h]
    \centering
    \begin{minipage}{0.48\textwidth}
    \includegraphics[width=\textwidth]{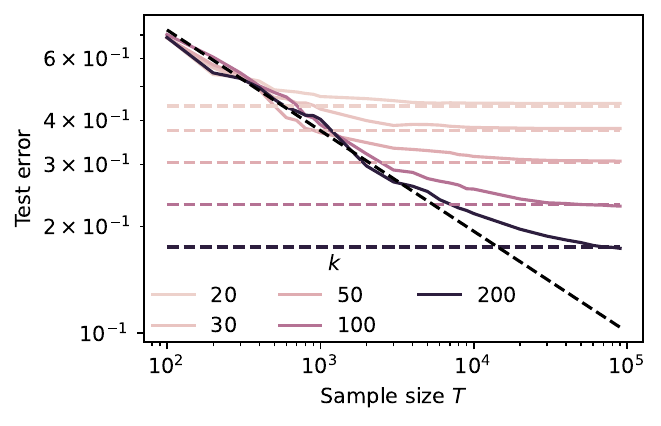}
    \vspace{-.5cm}
    \caption{\textbf{Model Collapse for Hutter++.} Empirical confirmation of Theorem \ref{thm:bigram3}. Here $p(j \mid i)$ is as in \eqref{eq:tailoutput}, with $\beta = 7/5$. The horizontal broken lines correspond to $k^{-\beta c}$ for different values of $k$, where $c:=\min(1-1/\beta,1/2)$. The diagonal broken line corresponds to $T^{-c}$ (classical error rate without cutoff).}
    \label{fig:bigram3}
    \end{minipage}
    \hfill
    \begin{minipage}{0.48\textwidth}
    \includegraphics[width=.9\linewidth]{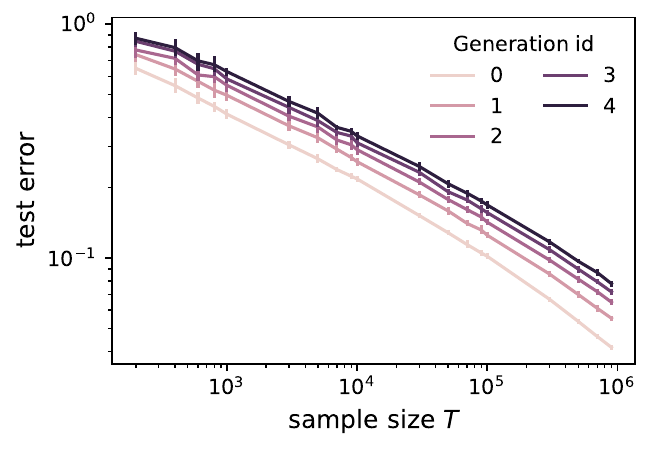}
    \vspace{-.5cm}
    \caption{\textbf{Hutter++ Model on Paired Bigram Data.} Empirical confirmation of Theorem \ref{thm:n_law} for probabilistic paired bigram data with $\beta=3/2$, $T_0 = T$ across all generations with no additional tail-cutting, regeneration for 9 times. The result verifies the model collapse across generation. }
    \label{fig:bigram_nocutoff}
    \end{minipage}
    \vspace{-0.5cm}
\end{figure}

\paragraph{Hutter++.} We now provide complementary illustrations of predictions made from the theory we have developed for the generalized Hutter models as in Equation \eqref{eq:tailoutput} in Section \ref{sec:bigram}, without departing from our theoretical assumptions. We also show how theory from the infinite memory model in Section \ref{sec:hutter} continues to hold in this bigram setting. Figure \ref{fig:bigram3} confirms the scaling law of Theorem \ref{thm:bigram3}.

In Figure \ref{fig:grokk} (middle) we have seen an illustration of the translated scaling curves under n-fold synthesized data in the Hutter LLM.
Figure \ref{fig:bigram_nocutoff} illustrates this phenomenon for the slightly more complex tailed bigram model.

\begin{figure}[htb]
    \centering
     \begin{minipage}{0.48\textwidth}
    \includegraphics[width=\linewidth]{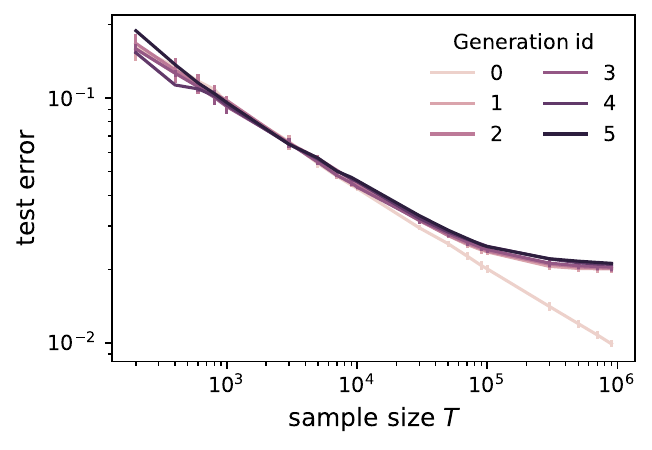}
    \caption{\textbf{Empirical Hutter LLM.} Bigram model with deterministic labeling function. Initial model trained on $T_0=100,000$ samples. It generates $T$ samples for Gen 1. Starting from Gen 2 models are trained on data generated by the most powerful model from the previous generation. $\beta=3/2$. In this setting, there is mild model collapse coming from the finite sample bias.}
    \label{fig:hutter_scale_n_new_topp100}
    \end{minipage}
    \hfill
     \begin{minipage}{0.48\textwidth}
      \includegraphics[width=\linewidth]{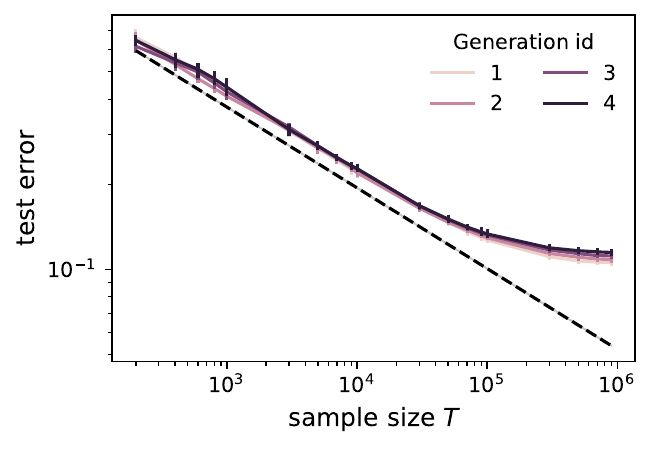}
    \caption{\textbf{Empirical Hutter++ Model.} Same setting as in Figure \ref{fig:top-p-95}. Initial model trained on $T_0=100,000$ samples. No top-$p^{inf}$ inference or temperature scaling is used. $\beta=3/2$. In this setting, there is mild model collapse coming from the finite sample bias as well. }
    \label{fig:no-topp}
     \end{minipage}
\end{figure}

\begin{figure}[htb]
    \centering
     \begin{minipage}{0.48\textwidth}
    \includegraphics[width=\linewidth]{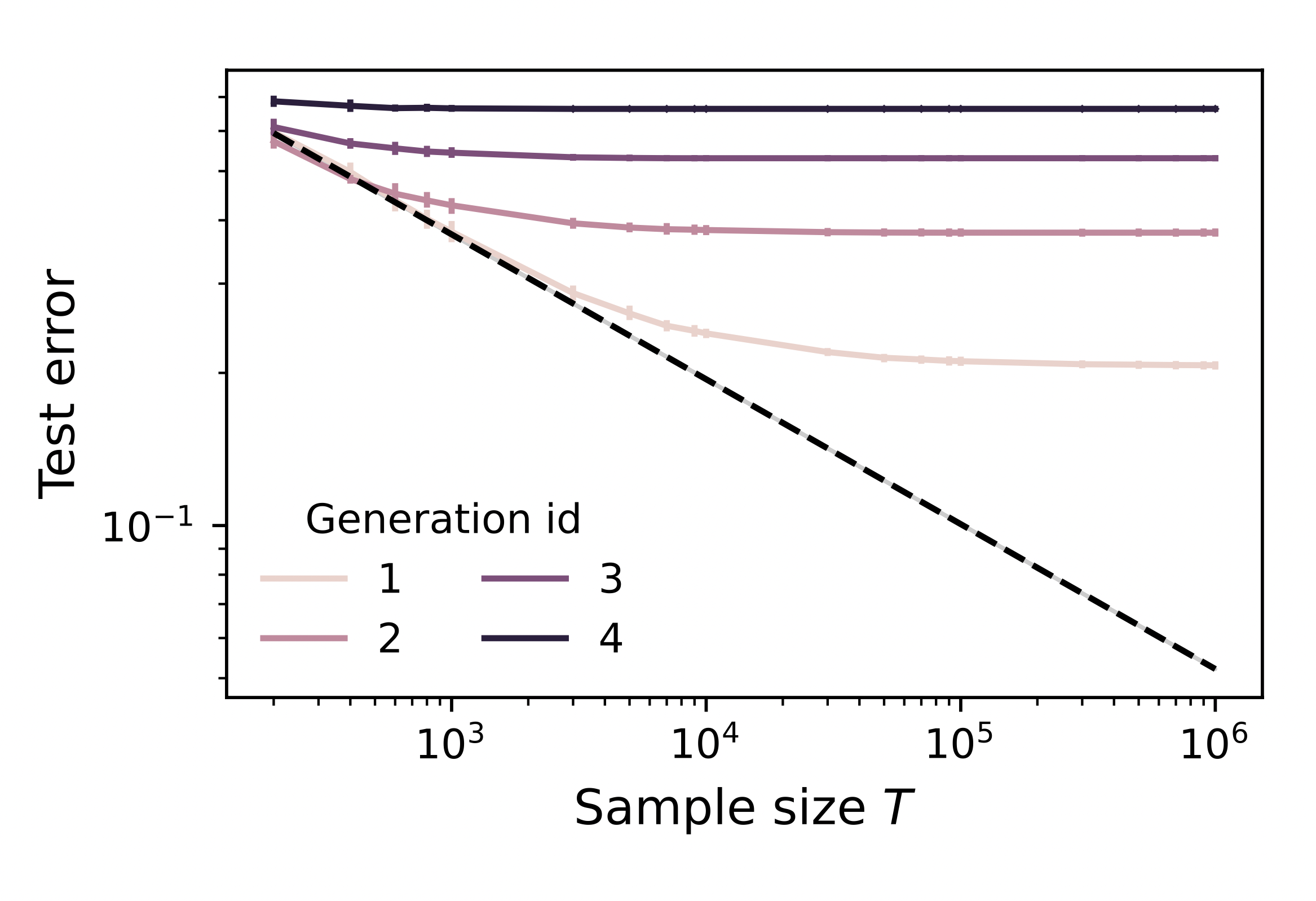}
    \caption{\textbf{Empirical Hutter++ Model.} Same setting as in Figure \ref{fig:no-topp} with top $p^{inf}=0.9$ synthesizing. No temperature scaling is used. $\beta=3/2$. Top-$p^{inf}$ selection significantly deteriorate the model collapse.}
    \label{fig:top-p-90}
    \end{minipage}
    \hfill\ 
    \begin{minipage}{0.48\textwidth}
    \includegraphics[width=\linewidth]{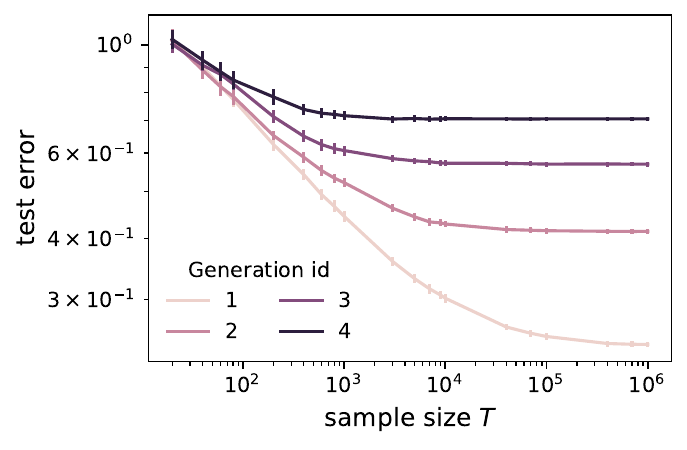}
    \caption{\textbf{Empirical Hutter++ Model.} Same setting as in Figure \ref{fig:no-topp} with temperature $\tau=$0.9 synthesizing. No top-$p^{inf}$ selection is used. $\beta=3/2$. Compared with Figure \ref{fig:no-topp}, temperature also create strong model collapse across multiple generation.}
    \label{fig:temp90}
    \end{minipage}
\end{figure}

Both Figures \ref{fig:grokk} (middle) and \ref{fig:bigram_nocutoff} illustrate the setting where each model consumes as much training data as its predecessor ($T_0=T$). We now relax the assumption that each successive model has strictly the same amount of training data as its predecessor. We assume that the generation 0 model is trained on $T_0$ (here, $T_0=100,000$) amount of original data to generate AI data for generation 1. All future generations, starting from generation 2, are trained on data generated by the most powerful model from the previous generation ($T=1,000,000$ data in this case). Figure \ref{fig:hutter_scale_n_new_topp100} (for Hutter LLM) and \ref{fig:no-topp} (for Hutter++ on paired bigram data) show the resulting scaling behavior. We take this setting even further by adding a top-$p^{inf}$ tail cutting mechanism and a temperature scaling mechanism for each synthetic data generation. Figure \ref{fig:top-p-95} cuts at $p=0.95$ and Figure \ref{fig:temp90} at temperature $0.9$.



\begin{figure}[htb]
    \centering
    \begin{minipage}{0.48\textwidth}
    \includegraphics[width=.8\linewidth]{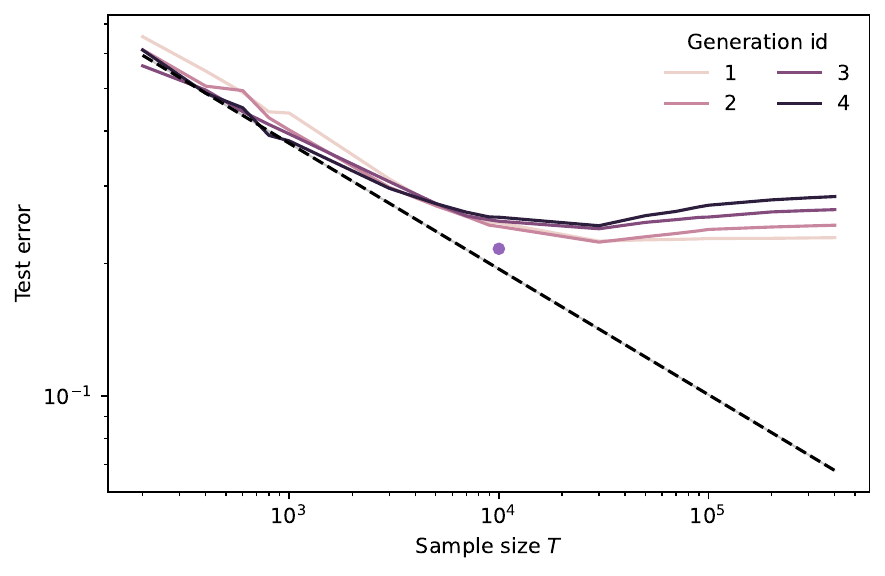}
    \caption{{\bf Empirical Hutter++ Model with Mixing.} The initial        ``clean" dataset comprises $T_0=10,000$ samples. For future generations, the largest model is used to synthesize data. For $T\leq 20,000$, training data is an equal mix of clean and generated data, for $T > 20,000$ all clean data is used; the remaining training data is synthetic (so the ratio of clean data diminishes). Top-$p^{inf}=$0.9, no temperature scaling, $\beta=3/2$. }
    \label{fig:topp95-mix}
    \end{minipage}
    \hfill
    \begin{minipage}{0.48\textwidth}
    \includegraphics[width=.8\linewidth]{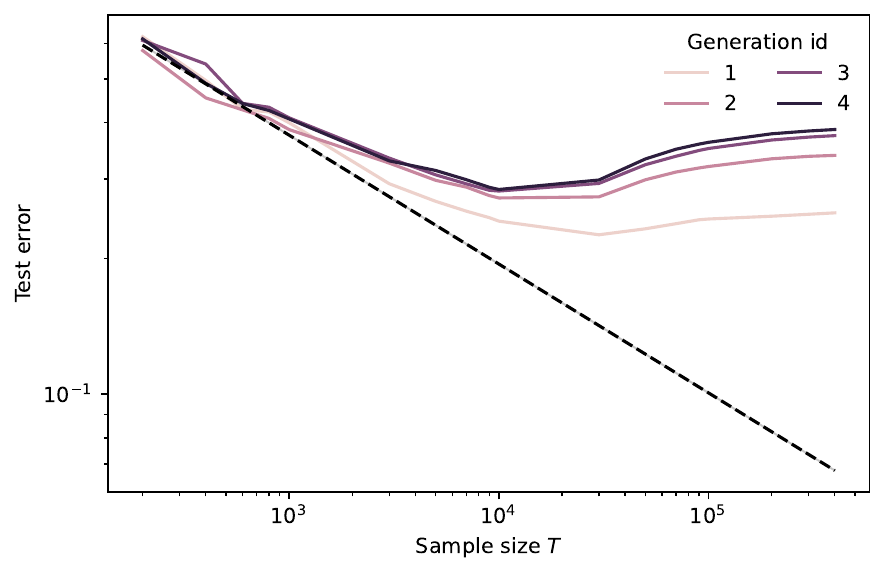}
    \caption{{\bf Empirical Hutter++ Model with Mixing.} Same setting as in Figure \ref{fig:topp95-mix} with top-$p^{inf}=0.9$, no temperature scaling, and $\beta=3/2$.}
    \label{fig:topp90-mix}
    \end{minipage}
\end{figure}

We now study mixing of clean and synthesized data in the bigram setting. Figures \ref{fig:topp95-mix} and \ref{fig:topp90-mix} add top-$p^{inf}$ tail-cutting when synthesizing, and start with $T_0=10,000$ original data samples, which are successively blended with synthesized data from the largest model. Note that in this setting we observe a reversion of scaling laws with increased AI data. This needs to be compared with the orange curve in Figure \ref{fig:mixbump} in the deterministic Hutter setting. The probabilistic nature of the bigram models leads to a new effect here.

\begin{figure}
    \centering
    \includegraphics[width=0.45\linewidth]{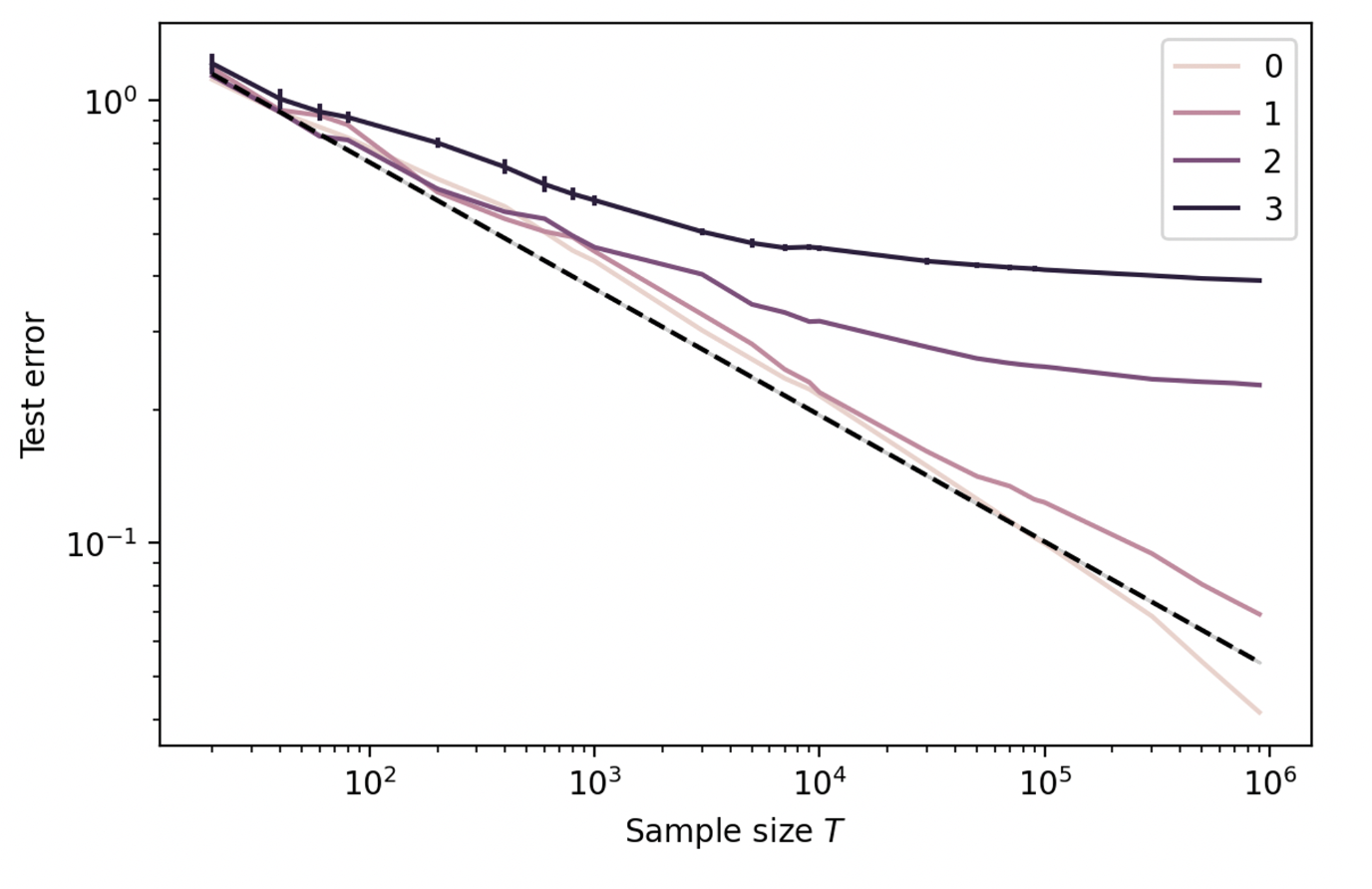}
    \caption{\textbf{S-shape ``Smoothed Grokking".} Bigram data with Hutter++ model, mixing clean data with AI generated data with ratio 50 to 50. The grokking line is smoothed in the probabilistic setting. Line 1, 2, 3 are generated by using 10,000, 1,000, and 100 data to train the generating model. Compared to Figure \ref{fig:topp95-mix}, we do not limit the number of accessible real data now. $\beta=3/2$.}
    \label{fig:s-shape}
\end{figure}

\section{Proofs for the infinite memory (Hutter) Model (Sections \ref{sec:hutter} and \ref{sec:grok})}\label{app:hutter}

 \subsection{Proof of Theorem \ref{thm:simple}}
 Observe that the model $\widehat f$ makes an error on $i$ if and only if the $i$th ``skill" never occurred in the training dataset $\mathcal D_T$, i.e either (1) $i \ge k + 1$, or (2) $1 \le i \le k$ and $i_t \ne i$ for all $t \in [T]$. We deduce that
\begin{eqnarray*}
\begin{split}
 E_{test} = \mathbb P_{i \sim p}(\widehat f(i) \ne y_i) &= \sum_{i \ge k+1}p_i + \sum_{1 \le i \le k}p_i(1-p_i)^T\\
 &\asymp k^{-(\beta-1)} + \sum_{1 \le i \le k}p_i e^{-p_i T},
 \end{split}
\end{eqnarray*}
where $c := 1-1/\beta \in (0,1)$, and we have used the elementary fact that $\sum_{i \ge k + 1}i^{-\beta} \asymp k^{-(\beta-1)}$ for large $k$. 
For the second sum, we will need the following lemma.
\begin{lemma}
The following identity holds
\begin{eqnarray}
T^c \sum_{i=1}^k p_i e^{-T p_i} \asymp \Gamma(c,Tk^{-\beta})-\Gamma(c,T) = O(1),
\end{eqnarray}
 where $\Gamma(s,x) := \int_x^\infty u^{s-1}e^{-u}\mathrm{d}u$ defines the incomplete gamma function. In particular, for $k=\infty$ and large $T$, it holds that $\sum_{i=1}^\infty p_i e^{-T p_i} \asymp T^{-c}$.
\label{lm:gamma}
\end{lemma}

\begin{proof}
Consider the function $h(z) := z e^{-Tz}$ for $z \in (0,1)$. Its derivative is $h'(z) = e^{-Tz}(1-Tz)$. Thus, $h$ is increasing on $(0, 1/T)$ and decreasing on $(1/T,\infty)$. Furthermore, note that $p_i \le 1/T$ iff $i \ge T^{1/\beta}$. We deduce that
$$
\sum_{i=1}^k p_i e^{-Tp_i} \asymp \int_1^k x^{-\beta} e^{-T x^{-\beta}}\mathrm{d}x.
$$

Under the change of variable $u=u(x):=Tx^{-\beta}$, we have $x=x(u)=(u/T)^{-1/\beta}$ and so $\mathrm{d}x = -(T^{1/\beta} u^{-1-1/\beta} / \beta)\mathrm{d}u$. Also $u(1)=T$ and $u(k) = Tk^{-\beta}$. We deduce that
\begin{align*}
 \sum_{i=1}^k p_i e^{-Tp_i} &\asymp \int_1^k x^{-\beta} e^{-T x^{-\beta}}\mathrm{d}x =  \int_{Tk^{-\beta}}^T (u/T) e^{-u}(T^{1/\beta} u^{-1-1/\beta} / \beta)\mathrm{d}u\\
 &\asymp T^{-(1-1/\beta)}\int_{Tk^{-\beta}}^T u^{-1/\beta} e^{-u}\mathrm{d}u\\
 &\asymp T^{-(1-1/\beta)}\left(\Gamma(1-1/\beta,Tk^{-\beta}) - \Gamma(1-1/\beta,T)\right)\\
 &= T^{-c}\left(\Gamma(c,Tk^{-\beta}) - \Gamma(c,T)\right),
 \end{align*}
 and we are done for the first part.

 For the second part, note that $\Gamma(c,T) = o(1)$ for large $T$ so that
 $$
 (\Gamma(c,Tk^{-\beta}) - \Gamma(c,T))|_{k=\infty} = \Gamma(c,0) - \Gamma(c,T) = \Theta(1) - o(1) = \Theta(1),
 $$
 from which the result follows. 
\end{proof}

 We now consider two separate cases for the relative scaling of $k$ and $T$.
 
 \paragraph{-- Case 1: $T \gtrsim k^\beta$.}
 Here, we have thanks to Lemma \ref{lm:gamma}
 \begin{eqnarray}
     \begin{split}
 E_{test} &\asymp k^{-(\beta-1)} + O(T^{-c}) \asymp k^{-(\beta-1)},
     \end{split}
 \end{eqnarray}
since $k^{-(\beta-1)} \gtrsim T^{-(\beta-1)/\beta} = T^{-c}$.

 \paragraph{-- Case 2: $1 \ll T \lesssim k^\beta$.} Here, thanks to Lemma \ref{lm:gamma} we have $\Gamma(c,T) = o(1)$ and $\Gamma(c,Tk^{-\beta}) = \Theta(1)$. We deduce that
 \begin{eqnarray}
 \begin{split}
     E_{test} &\asymp k^{-(\beta-1)} + T^{-c}\left(\Gamma(c,Tk^{-\beta}) - \Gamma(c,T)\right) \asymp k^{-(\beta-1)} + T^{-c} \asymp T^{-c},
     \end{split}
 \end{eqnarray}
 since $k^{-(\beta-1)} \lesssim T^{-(\beta-1)/\beta} = T^{-c}$.
Putting things together then gives the claimed result. \qed

\subsection{Proof of Corollary \ref{cor:narrowtail}}
Indeed, let $p_i \propto i^{-\beta}$ and $(p_{AI})_i = q_i \propto i^{-\beta'}$. Then,
\begin{eqnarray}
    E_{test} \asymp \sum_i p_i (1-q_i)^T \asymp \sum_i p_i e^{-q_i T} \asymp \int_1^\infty x^{-\beta} e^{-x^{-\beta'} T}\mathrm{d}x.
\end{eqnarray}
Setting $u = x^{-\beta'} T$ gives $x=T^{1/\beta'}u^{-1/\beta'}$, and so $\mathrm{d}x = -(T^{1/\beta'}/\beta') u^{-(1+1/\beta')}\mathrm{d}u$. We deduce that
\begin{eqnarray*}
\begin{split}
    E_{test} &\asymp T^{-(\beta-1)/\beta'}\int_1^T u^{\beta/\beta'} u^{-(1+1/\beta')} e^{-u}\mathrm{d}u = T^{-(\beta-1)/\beta'}\int_1^T u^{(\beta-1)/\beta' - 1}e^{-u}\mathrm{d}u\\
    &\asymp T^{-c} \Gamma(c,T) = T^{-c}(1+o(1)),\text{ with }c := (\beta-1)/\beta'.
    \end{split}
\end{eqnarray*}
That is, $E_{test} \asymp T^{-c}$ as claimed. \qed

\subsection{Proof of Theorem \ref{thm:grokk} and Corollary \ref{cor:bump}}
Suppose that of $T$ samples available for training our model, $\pi T$ are samples from the true distribution $p=Zipf(\beta)$ and $(1-\pi) T$ are from AI data distribution $p'$ which is a version of $p$ with its tail chopped off at rank $k$, i.e such that $p'_i \propto p_i 1[i \le k]$. Thus the dataset is drawn from the distribution given by $q_i = \pi p_i + (1-\pi) p'_i$. Test error of a Hutter LLM then writes
\begin{eqnarray}
\begin{split}
    E_{test} &= \sum_{i \ge1} p_i(1-q_i)^T = \sum_{1 \le i \le k} p_i(1-p_i)^T + \sum_{i \ge k+1} p_i(1-\pi p_i)^T\\
    &\asymp \sum_{1 \le i \le k} p_i e^{-p_i T} + \sum_{i \ge k + 1} p_ie^{-\pi p_i T}.
    \label{eq:Epi}
    \end{split}
\end{eqnarray}
Now, thanks to Lemma \ref{lm:gamma}, it is clear that for any integers $1 \le r<R\le \infty$ and large $z$, one has
\begin{eqnarray}
\label{eq:toptop}
\sum_{r \le i \le R}p_i e^{-p_i z} \asymp z^{-c}\left(\Gamma(c,z R^{-\beta}) - \Gamma(c,z r^{-\beta})\right),
\end{eqnarray}
where $c = 1-1/\beta \in (0,1)$ and $\Gamma$ is the (upper) incomplete gamma function. Applying \eqref{eq:toptop} with $(r,k,z) = (1,k,T)$ gives
\begin{eqnarray}
    T^c\sum_{1 \le i \le k}p_i e^{-p_i T} \asymp \Gamma(c,Tk^{-\beta})-\Gamma(c,T)
    = \begin{cases}
        \Theta(1)-o(1) = \Theta(1),&\mbox{ if }1 \ll T \lesssim k^\beta,\\
        o(1) - o(1) = o(1),&\mbox{ if }T \gtrsim k^\beta \gg 1.
    \end{cases}
\end{eqnarray}
On the other hand, applying \eqref{eq:toptop} with $(r,k,z) = (k+1,\infty,\pi T)$ and assuming $\pi = \Theta(1)$ gives
\begin{eqnarray}
    \sum_{i \ge k + 1}p_i e^{-\pi p_i T} \asymp (\pi T)^{-c}\gamma(c,\pi T(k+1)^{-\beta}) \asymp \begin{cases}
        (\pi T)^{-c},&\mbox{ if }\pi T \gtrsim k^\beta \gg 1,\\
        (k+1)^{-\beta c} \asymp k^{-\beta c},&\mbox{ if }k^\beta \gg \pi T. 
    \end{cases}
\end{eqnarray}
Putting things together gives the result. \qed

Recall that \citet{bertrand2023stability} also formally study such mixtures for iterative retraining. In their setting, they show the existence of fixed points in the mixture proportion that delineates the region of model collapse. These results are complimentary and not contradictory to ours: they combine mixing, large number of iteration, and data-decay, thus studying a combination of effects (under different theoretical conditions, not focusing on scaling laws) that our preceding theorems address separately.

\subsection{Grokking for Tail Narrowing}\label{app:groknarrow}

\begin{theorem}[Grokking with Tail Narrowing]
Consider a sample of size $T$ of which a proportion $\pi$ comes from the true distribution $p=Zip(\beta)$ and the remainder comes from a version $p' = Zip(\beta')$. We have the following scaling law for the Hutter LLM,
\begin{eqnarray}
    E_{test} \asymp (\pi T)^{-c} + ((1-\pi) T^{-c'}),
\end{eqnarray}
where $c := (\beta-1)/\beta$ and $c' := (\beta-1)/\beta'$.

Define $\overline T := (\pi / (1-\pi))^{-a}$, where $a := s / (1-s)$, and $s := \beta/\beta'$. Then,

(A) \textbf{Early-Stage Dynamics.} For $T \lesssim \overline T$, it holds that $E_{test} \asymp ((1-\pi)T)^{-c'}$. Thus,  if $\beta' > \beta$, the money spent on acquiring some clean data is not amortized!

(B) \textbf{Later-Stage Dynamics.} As soon as $ T \gtrsim \overline T $, it holds that $E_{test} \asymp (\pi T)^{-c}$. Similarly, we recover the unpolluted  sample-size law scaling $T^{-c}$. For fixed $T$ and tunable $\pi$, this error rate scales like $\pi^{-c}$.
\label{thm:grokk2}
\end{theorem}

\paragraph{Proof.} Let $q$ be the mixture of $p$ and $p'$. We prove the result for $\beta' \ge \beta$; the case $\beta' \le \beta$ is analogous. So, one may write
\begin{eqnarray}
    E_{test} = \sum_{i \ge 1} p_i(1-q_i)^T \asymp \sum_{i \ge 1} p_i e^{-\pi i^{-\beta} + (1-\pi) i^{-\beta'}} \asymp \sum_{1 \le i \le \overline T^{1/\beta}} p_i e^{-\pi i^{-\beta}} + \sum_{i \ge \overline T^{1/\beta}} p_i e^{-(1-\pi) i^{-\beta'}},
\end{eqnarray}
where we have used the fact that $(1-\pi) i^{-\beta'} \ge \pi i^{-\beta}$ iff $i \le (\pi/(1-\pi))^{-1/(\beta'-\beta)} = \overline T^{1/\beta}$. The result then follows from \eqref{eq:toptop}.\qed

\begin{remark}
Let us conclude by saying that clean data always helps, since $E_{test}$ is decreasing function of $\pi$. Indeed, from \eqref{eq:Epi}, the derivative w.r.t $\pi$ is $E_{test}'(\pi) = -T\sum_{i \ge k+1} p_i^2 (1-\pi p_i)^{T-1} \le 0$.
\end{remark}

\subsection{An interesting detour: Grokking for Fixed-size AI Dataset.}\label{app:limitedAI}
\begin{figure}[htb]
    \centering
    \includegraphics[width=0.45\linewidth]{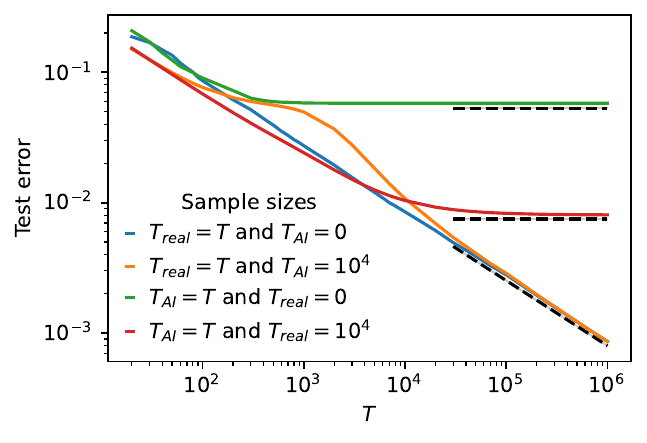}
    \vspace{-.5cm}
    \caption{{\bf Hutter LLM.} true distribution of the data is Zipf with exponent $\beta=2$. Here, the scalable resource is either clean data or AI data-generated data, corresponding to a version of real data with its tail cut at rank $k$ (here we use $k=10$). We either mix with a fixed amount (here $T' = 10^4$ samples) of the other resource, or we don't mix at all. Then we scale up the scalable resource by cranking up $T$. As predicted by Corollary \ref{cor:bump}, the orange curve always grokks: AI synthesized data is helpful in the regime where real data is scarce; once more of real data becomes available the model grokks for a while and then forgets the AI synthesized data. Note that the green curve (only AI data) and red curve (AI + real data) don't grokk because the optional resource (real data) is not being scaled; if it is also scaled, then green and red will provably grokk (as in Figure \ref{fig:grokk}). The diagonal broken line corresponds to the standard Hutter scaling law $E_{test} \asymp T^{-c}$, where $c:=1-1/\beta$. The horizontal broken lines correspond to $E_{test} \asymp k^{-(\beta-1)}$ and $E_{test} \asymp T'^{-c}$, both predicted by Theorem \ref{thm:simple}. }
    \label{fig:mixbump}
    \vspace{-.5cm}
\end{figure}

Now consider the scenario where the AI synthesized dataset has fixed size $T_{AI}$ (e.g a frozen chunk of the web), while the clean dataset size is a scalable parameter $T_{real}$. Taking $T=T_{real}+T_{AI}$ and $\pi=T_{real}/T$, we have the following corollary of Theorem \ref{thm:grokk}, which includes Corrolary \ref{cor:bump}. 
\begin{corollary}
We have the following.

\textbf{(A) Early-Stage Dynamics.} For $T_{real} \ll k^\beta$, it holds that
\begin{eqnarray}
    E_{test} \asymp (T_{real} + T_{AI})^{-(1-1/\beta)} + k^{-(\beta-1)}
\end{eqnarray}
\textbf{(B) Later-Stage Dynamics.} As soon as $T_{real} \ge Ck^{\beta}$ (where $C$ is an absolute constant), it holds that
\begin{eqnarray}
    E_{test} \asymp T_{real}^{-(1-1/\beta)}.
\end{eqnarray}
\label{cor:bump2}
\end{corollary}
\vspace{-.5cm}
As mentioned in Section \ref{sec:grok}, AI synthesized data is helpful in the regime where real data is scarce. Once more of real data becomes available the model grokks for a while and then forgets the AI synthesized data to recover the normal scaling law w.r.t $T_{real}$. Figure \ref{fig:mixbump} gives an illustration of this phenomenon in various settings.

\subsection{Proof of Theorem \ref{thm:annealed}}
Note that explicitly,
\begin{eqnarray}
    \pi_i \asymp \begin{cases}
    N^\alpha p_i,&\mbox{ if }i \ge N,\\
    0,&\mbox{ else,}
    \end{cases}
\end{eqnarray}
where $\alpha := \beta-1$. This is because the normalization constant is $\sum_{i \ge N} p_i = \sum_{i \ge N} i^{-\beta} \asymp N^{-\alpha}$.
 Now, mix this distribution with $q$ with equal weights $1/2$, to obtain a new distribution
\begin{eqnarray}
\begin{split}
    q'_i = q_i/2 + \pi_i/2 &= \begin{cases}
        q_i/2,&\mbox{ if }i \le k,\\
        \pi_i/2,&\mbox{ if }k \ge N,\\
        0,&\mbox{ otherwise}
    \end{cases}\\
    &\asymp \begin{cases}
        p_i,&\mbox{ if }i \le k,\\
        N^\alpha p_i,&\mbox{ if }k \ge N,\\
        0,&\mbox{ otherwise,}
    \end{cases}
\end{split}
\end{eqnarray}
For simplicity, assume $N \ge k + 1$ (otherwise, we have all of $p$). Build a "Hutter" LLM from an iid sample of size $T$ from this distribution (this is equivalent to mixing $T$ samples from $q$ and $T$ samples from $\pi$.
Then, it is easy to see that the test error is given by
\begin{eqnarray}
    E_{test} = \sum_{i \ge 1}p_i(1-q_i')^T \asymp \sum_{1 \le i \le k} p_i(1-p_i)^T + \\ \nonumber \sum_{k + 1 \le i \le N - 1} p_i + \sum_{i \ge N} p_i(1-N^{\alpha}p_i)^T.
\end{eqnarray}
Thanks to previous computations, we know that for large $k$, $N$, and $T$
\begin{itemize}
    \item The first sum is of order $T^{-c}\left(\Gamma(c,Tk^{-\beta}) - \Gamma(c,T)\right) =O(T^{-c})$.
    \item The third sum is of order $T^{-c}\left(\Gamma(c,0) - \Gamma(c,TN^\alpha N^{-\beta})\right) = T^{-c}\left(\Gamma(c,0) - \Gamma(c,TN)\right) \asymp T^{-c}$.
    \item The second sum is of order $k^{-\alpha} -N^{-\alpha} =((\frac{N}{k})^\alpha - 1) N^{-\alpha}$, where $\alpha := \beta-1$.
\end{itemize}
We deduce that
\begin{eqnarray}
\begin{split}
    E_{test} &\asymp T^{-c} + \left(\left(\frac{N}{k}\right)^\alpha-1\right)N^{-\alpha},\text{ for large }k,N,T,
    \end{split}
\end{eqnarray}
and the result follows.
\qed

\section{Proofs for the Tailed Bigram Model (Section \ref{sec:bigram})}\label{app:bigram}

\subsection{Warm-up: Revisiting the Classical Hutter Setup}\label{app:gethutterback}

As a sanity check, with the framework of Equation \eqref{eq:TVloss}, let us momentarily consider the non-autoregressive setup where $p(\cdot \mid i) = \delta_{y_i}$ for all $i$, as in classical Hutter. Then, an easy computation shows that
$$
TV(q_T(\cdot\mid i),p(\cdot\mid i)) = 1-q_T(y_i \mid i) + \sum_{j \ne y_i} q_T(j \mid i) = 2(1-q_T(y_i \mid i)).
$$

Now, by construction, $q_T(y_i \mid i) = 1[i \in \mathcal D_T]$. Thus,
$$
\mathbb E\, [1-q_T(y_i \mid i)] = \mathbb P(i \not \in \mathcal D_T) = (1-p_i)^T.
$$
We deduce that
$$
\mathbb E\,[TV(q_T(\cdot\mid i),p(\cdot\mid i))] = 2(1-p_i)^T.
$$
Therefore,
\begin{eqnarray}
    E_{test} = \sum_i p_i \mathbb E\,[TV(q_T(\cdot\mid i),p(\cdot\mid i))] \\ \nonumber = 2\sum_i p_i(1-p_i)^T \asymp T^{-(1-1/\beta)},
\end{eqnarray}
and we recover the classical Hutter result! Thus, our test metric defined in \eqref{eq:TVloss} is pointing in the right direction, conceptually.

\subsection{Proof of Theorem \ref{thm:bigram1}}
The proof will be based on the results of \cite{Berend2012OnTC}. (\ElvisIssue{Need to check this; maybe martingale arguments are needed here!})

 \paragraph{Upper-Bound.}
 Observe that for any choice of mappings $\pi_1,\pi_2,\ldots$, we have 
\begin{eqnarray*}
    \begin{split}
a_T(i) &:= \sum_{j\, \mid \,p(j \mid i) \le 1/n_T(i)} p(j \mid i) \asymp \sum_{j\, \mid \,\pi_i(j) \ge n_T(i)^{1/\beta}} \pi_i(j)^{-\beta} \le \sum_{k\, \mid \, k \ge n_T(i)^{1/\beta}} k^{-\beta} \asymp n_T(i)^{-(1-1/\beta)}\\
b_T(i) &:= n_T(i)^{-1/2}\sum_{j\, \mid \,p(j \mid i) \ge 1/n_T(i)} \sqrt{p(j \mid i)} \asymp n_T(i)^{-1/2}\sum_{j\, \mid \, \pi_i(j) \le n_T(i)^{1/\beta}}\pi_i(j)^{-\beta/2}\\
&\lesssim n_T(i)^{-1/2}\sum_{k\, \mid \, k \le n_T(i)^{1/\beta}}k^{-\beta/2} \asymp n_T(i)^{-c}.
    \end{split}
\end{eqnarray*}
We deduce that
$c_T(i) := a_T(i) + b_T(i) \lesssim n_T(i)^{-c}$ for any $i$.  Importantly, the hidden constants don't depend on $i$.  Therefore, thanks to [Lemma 9] \cite{Berend2012OnTC}, we have
\begin{eqnarray}
\label{eq:tvub}
\begin{split}
    E_{test} \le \sum_i p_i \mathbb E\,[c_T(i)] &\lesssim \sum_i p_i \mathbb E\,[n_T(i)^{-c}] \overset{(*)}{\le} \sum_i p_i (\mathbb E\,[n_T(i)])^{-c} = \sum_i p_i (T p_i)^{-c} = T^{-c}\sum_i p_i^{1-c}\\
    &\lesssim T^{-c},
    \end{split}
\end{eqnarray}
where we have used Jensen's inequality in (*), since the function $x \mapsto x^{-c}$ is concave.

\paragraph{Lower-Bound.} WLOG\footnote{A summable series of nonnegative numbers (like in $a_T(i)$ and $b_T(i)$) can be reordered without changing the value.} consider the following specific choice of permutations defined by $\pi_i(j) = j$ (i.e doesn't depend on $i$). Then, 
\begin{align*}
a_T(i) &= \sum_{j \ge n_T(i)^{1/\beta}} j^{-\beta} \asymp n_T(i)^{-(1-1/\beta)},
\\
b_T(i) &= n_T(i)^{-1/2}\sum_{j \le n_T(i)^{1/\beta}} j^{-\beta} \asymp n_T(i)^{-c}.
\end{align*}
Thanks to the definition of $E_{test}$ and [Proposition 5] \cite{Berend2012OnTC}, we deduce that if $\beta \in (1,2)$, then
\begin{eqnarray}
E_{test} \ge \sum_i p_i \mathbb E\,[(a_T(i) + b_T(i) - n_T(i)^{-1/2})] \asymp \sum_i p_i \mathbb E\,[n_T(i)^{-c}- n_T(i)^{-1/2})] \asymp \sum_i p_i \mathbb E\,[n_T(i)^{-c}],
\label{eq:tvlb}
\end{eqnarray}
i.e $E_{test} \gtrsim \sum_i p_i \mathbb E\,[n_T(i)^{-c}]$. Now, since $n_T(i) \sim Bin(T,p_i)$, standard Binomial concentration arguments tell us that $n_T(i) \le 1.5 T p_i$ w.p $1-e^{-Cp_i T}$, where $C$ is an absolute constant. We deduce that
\begin{eqnarray*}
\begin{split}
    E_{test} &\gtrsim \sum_i p_i (1.5 T p_i)^{-c}(1-e^{-Cp_iT})\asymp T^{-c} \sum_i p_i^{1-c} - T^{-c} \underbrace{\sum_i p_i^{1-c} e^{-Cp_iT}}_{o(1)} \asymp T^{-c},
    \end{split}
\end{eqnarray*}
which completes the proof. \qed

\subsection{Proof of Theorem \ref{thm:bigram3}}
It suffices to replace $n_T(i)$ in \eqref{eq:tvub} and \eqref{eq:tvlb} of the proof of Theorem \ref{thm:bigram1} with $n_T(i) \land k^\beta$, and use the elementary fact that $(n_T(i) \land k^\beta)^{-c} = n_T(i)^{-c} \lor k^{-\beta c} \asymp n_T(i)^{-c} + k^{-\beta c}$. The rest of the proof proceeds as that of Theorem \ref{thm:bigram1}. 

\subsection{Extensions}

Note that the above setup can be extended to the following
$$
p(j \mid i) = \rho(\pi_i(j)),
$$
where $\rho$ is a distribution on $\mathbb N_*$. In particular, taking $\rho(z) \propto z^{-\beta}$, recovers the setup considered above. It is clear that mechanics of the proof of Theorem \ref{thm:bigram1} should be applicable here, leading to scaling laws which depend explicitly on $\rho$.

\section{Proof and Illustration of Triplet Scaling Law (Theorem \ref{thm:triplet}) }\label{app:triplet}
For any $i$, on average it takes $1/p_i$ iid samples from $p$ to see the context $i$ at least once.
The effect of tail-cutting at rank $k$ is effectively to replace the sample size $T$ by $\min(T,T_k)$, where $T_k = \max \{1/p_i \mid i \in [k]\}$.
In the case where $p = Zipf(\beta)$, we have $T_k = 1/p_k \asymp k^{\beta}$. On other hand the model \eqref{eq:eva} proposed in \cite{cabannes2023scaling} on Zipf data, the test error writes
\begin{eqnarray}
    E_{test} \asymp T^{-c} + d^{-c_q},
\end{eqnarray}
where $c := 1-1/\beta \in (0,1)$ and the exponent $c_q \in (0,\infty)$ depends on $\beta$ and the algorithm $q$ used to update the embeddings in the memory matrix $M_T$ in \eqref{eq:eva}.
We deduce that tail-cutting at rank $k$ changes the test error to
$$
E_{test} \asymp \min(T,T_k)^{-c} + d^{-c_q} \asymp T^{-c} + k^{-\beta c} + d^{-c_q},
$$
as claimed. \qed

\begin{figure*}
    \centering
    \includegraphics[width=1\textwidth]{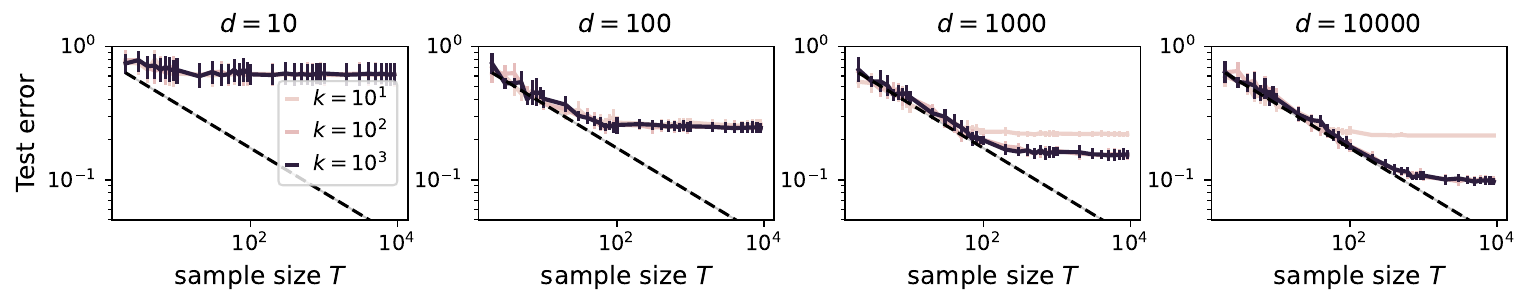}
    
    \includegraphics[width=1\textwidth]{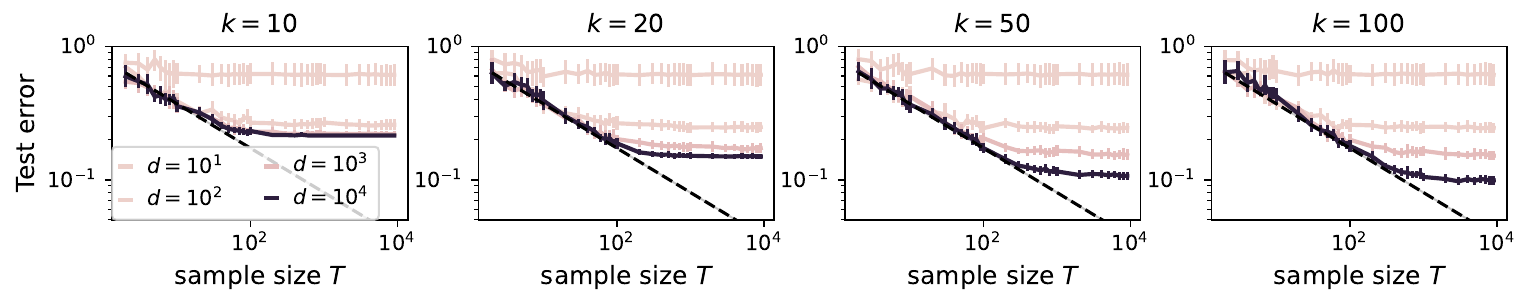}
    \caption{{\bf Capacity-Limited Memory Models.} Empirical confirmation of the Triplet Scaling Law established in Theorem \ref{thm:triplet}}
    \label{fig:tripletlaw}
\end{figure*}

Figure \ref{fig:tripletlaw} confirms the Triplet Scaling Law.

\clearpage

\section{Details and Results from the Autoregressive Bigram model with Perplexity}\label{app:perplexity}

\begin{figure}[htb]
    \centering
    \begin{minipage}{0.48\textwidth}
    \includegraphics[width=\linewidth]{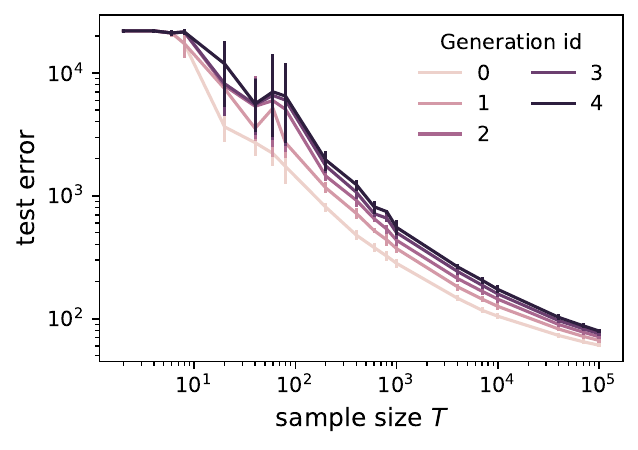}
    \caption{{\bf Autoregressive Bigram Model with Perplexity Loss.} Empirical confirmation of Theorem \ref{thm:n_law} for autoregressive data with top-$p^{inf}=$1, Temperature $\tau$ 1. Each sequence data have length 100. Same setting as Figure \ref{fig:bigram_nocutoff}. $\beta=3/2$. }
    \label{fig:seq_bigram100_1}
    \end{minipage}
    \hfill
    \begin{minipage}{0.48\textwidth}
    \includegraphics[width=\linewidth]{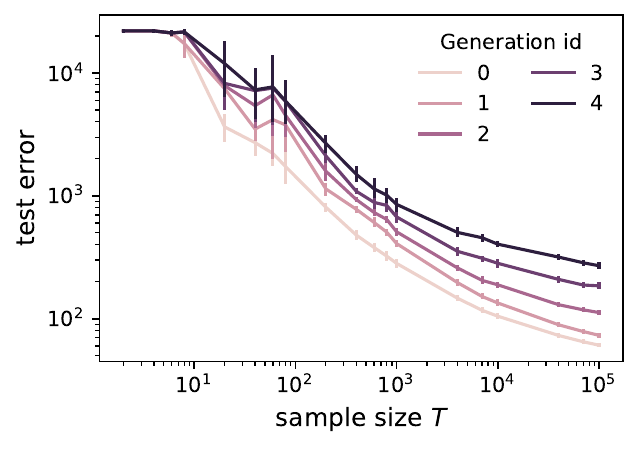}
    \caption{{\bf Autoregressive Bigram Model with Perplexity Loss.} Empirical confirmation of Theorem \ref{thm:n_law} for autoregressive data with top-$p^{inf}=$0.9, Temperature $\tau$ 1. Each sequence data have length 100. $\beta=3/2$.}
    \label{fig:seq_bigram90_1}
    \end{minipage}
\end{figure}

\begin{figure}[htb]
    \centering
    \begin{minipage}{0.48\textwidth}
    \includegraphics[width=\linewidth]{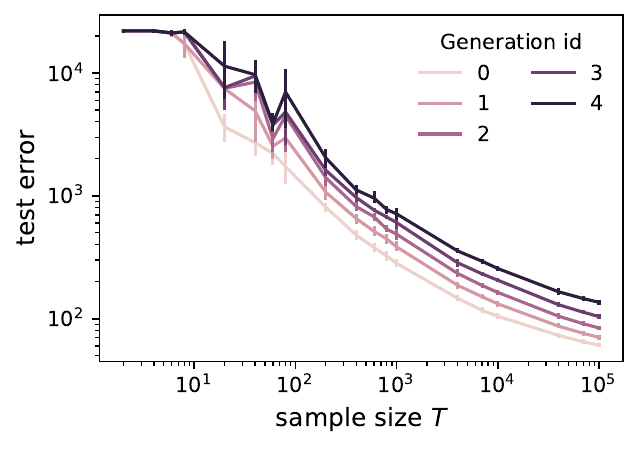}
    \caption{{\bf Autoregressive Bigram Model with Perplexity Loss.} Empirical confirmation of Theorem \ref{thm:n_law} for autoregressive data with top-$p^{inf}=$1, Temperature $\tau=$0.9. Each sequence data have length 100. Same setting as Figure \ref{fig:bigram_nocutoff}. $\beta=3/2$.}
    \label{fig:seq_bigram100_09}
    \end{minipage}
    \hfill
    \begin{minipage}{0.48\textwidth}
    \includegraphics[width=\linewidth]{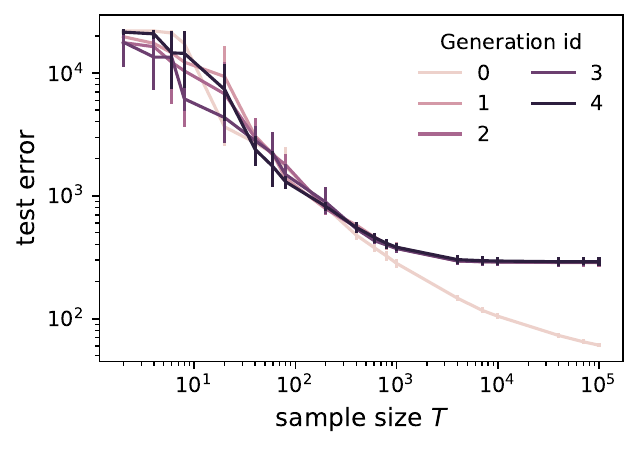}
    \caption{{\bf Autoregressive Bigram Model with Perplexity Loss.} Each sequence data have length 100. Initial model trained on $T_0=10,000$ samples. It generates $T$ samples for Gen 1. Starting from Gen 2 models are trained on data generated by the most powerful model from the previous generation. Top-$p^{inf}=$1, temperature $\tau=$1, $\beta=3/2$.}
    \label{fig:seq_bigram_new_100_1}
    \end{minipage}
\end{figure}

\begin{figure}[htb]
    \centering
    \begin{minipage}{0.48\textwidth}
    \includegraphics[width=\linewidth]{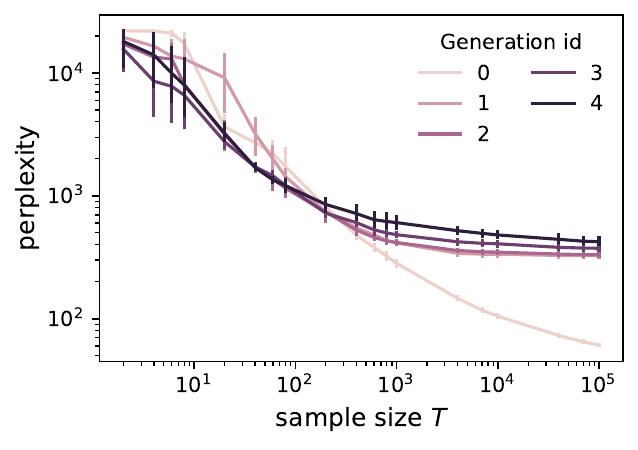}
    \caption{{\bf Autoregressive Bigram Model with Perplexity Loss.} Each sequence data have length 100. Same setting as Figure \ref{fig:seq_bigram_new_100_1}. Top-$p^{inf}=$0.9, temperature $\tau=$1, $\beta=3/2$.}
    \label{fig:seq_bigram_new_90_1}
    \end{minipage}
    \hfill
    \begin{minipage}{0.48\textwidth}
    \includegraphics[width=\linewidth]{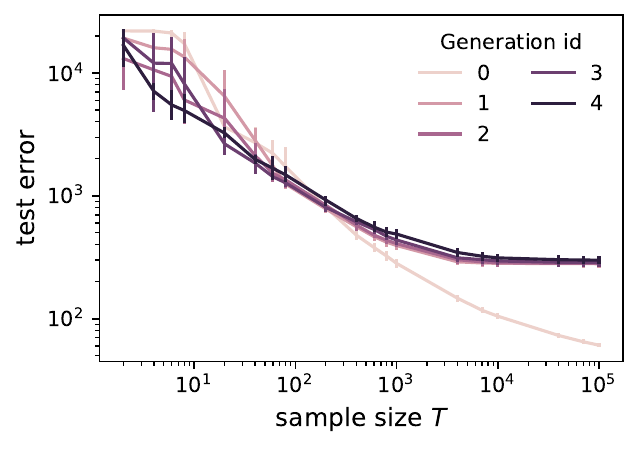}
    \caption{{\bf Autoregressive Bigram Model with Perplexity Loss.} Each sequence data have length 100. Same setting as Figure \ref{fig:seq_bigram_new_100_1}. Top-$p^{inf}=$1, temperature $\tau=$0.9, $\beta=3/2$.}
    \label{fig:seq_bigram_new_100_90}
    \end{minipage}
\end{figure}

We showcase experiments in the autoregressive bigram model with perplexity loss. We generate sequences of length $100$.  Figures \ref{fig:seq_bigram100_1}, Figure \ref{fig:seq_bigram90_1} and \ref{fig:seq_bigram100_09} aim to reproduce the ''paired bigram" Figure \ref{fig:bigram_nocutoff} in this setting, adding a top $p^{inf}$ mechanism and a temperature mechanism. Figure \ref{fig:seq_bigram_new_90_1}, Figure \ref{fig:seq_bigram_new_90_1} and Figure \ref{fig:seq_bigram_new_100_90} regenerates the setting of Figure \ref{fig:top-p-95} with the same top $p^{inf}$ and temperature.

\clearpage

\section{Details and Results on Transformer Arithmetic Experiments}\label{app:gcd}

\citet{charton2023transformers} trains sequence-to-sequence transformers to predict the greatest common divisor (GCD) of two positive integers, encoded as sequences of digits in some base $B$. He observes that model predictions are deterministic: for any pair $(a,b)$ with GCD $k$, the model predicts a single value $f(k)$. Predictions are correct (i.e. $f(k)=k$) when the GCD is a product of divisors of the base, or of small primes. In all other case, the model prediction is the largest correct prediction (i.e. $l$ such that $f(l)=l$) that divides $k$. The list of correct predictions $\mathcal L$ varies with the encoding base $B$. For instance, for $B=10$, after 300 million examples, the model correctly predicts $\mathcal L = \{1,2,4,5,8,10,16,20,25,40,50,80,100... \}$, the GCD of $20$ and $30$ will be correctly predicted as $10$, but the GCD of $210$ and $140$ will be incorrectly predicted as $10$ (instead of $70$). 

We use these models to generate ``dirty'' training data $\mathcal D(B)$: uniformly sampled pairs of integers $(a,b)$ and their (sometimes incorrect) pseudo-GCD, as generated by a trained transformer using base $B$. Note: this dataset can be as large as we want. We also create a correct training dataset $\mathcal C(B)$, by sampling pairs $(a,b)$ and their correct GCD.

In these experiments, we train models on $\mathcal D(B)$ and $\mathcal C(B)$, for different values of $B$. Our goal is to determine whether extensive training on ``dirty'' data impacts model accuracy.

We focus on $6$ bases: $B= 10, 420, 1000, 2017, 2023$ and $4913$, after training transformers (on correct GCD) over about $300$ millions pairs of integers between one and one million, we achieve the performances listed in Table~\ref{tab:base_results}. There, accuracy stands for the proportion of random uniform pairs $(a,b)$ that the model can predict correctly, correct GCD is the number of GCD under $100$ that the model correctly predicts (i.e. $k$ such that $f(k)=k$), and correct model predictions are the products of numbers in the associated sets. These models are used to generate $\mathcal D(B)$.

In these experiments, all models have four layers, 512 dimensions and 8 attention heads. We consider two architectures: an encoder-only model (17.2M parameters), and an encoder-decoder model (38.7M parameters). The encoder-only model has $2.25$ times less parameters, trains twice as fast, and incurs no performance penalty.

\begin{table}[h]
    \small
    \centering
    \caption{\small \textbf{Initial performances.} $4$-layer transformers trained to predict GCD, on 300 million examples. Our {\em test} set only contains GCD up to $100$, and accuracy is computed on a reweighted test  with equal occurance of each GCD. Thus, the Correct GCD lists all those that can be formed from the correct predictions by forming products across the sets (within the first 100 GCD). We freeze the 0th generation model at this stage and use its prediction to generate synthetic data. For each GCD outside the set of its correct predictions, the model will predict the largest GCD it has learned that divides the ground truth.}
    \begin{tabular}{c|ccl}
    \toprule
       Base  & Accuracy & Correct GCD & Correct predictions \\
       \midrule
       10 & 85 & 13 & \{1,2,4,8,16\} \{1,5,25\}  \\
       420 & 97 & 38 & \{1,2,4,8,16\}\{1,3,9\}\{1,5,25\}\{1,7\}\\
       1000 & 94 & 22 & \{1,2,4,8,16\} \{1,5,25\}\{1,3\} \\
       2017 & 85 & 4 & \{1,2\}\{1,3\} \\
       2023 & 91 & 16 & \{1,2,4\}\{1,3\}\{1,7\}\{1,17\} \\
       4913 & 93 & 17 &  \{1,2,4\}\{1,3\}\{1,5\}\{1,17\}\\
       \bottomrule
    \end{tabular}
    \label{tab:base_results}
\end{table}

We then train new models (with the same architecture) to predict GCD, from AI data (generated by the above model), and compare to training with correct data -- from correct computation of the GCD. When trained on small number of examples (less than $100$ million), models learning from AI data achieve better accuracy (Table~\ref{tab:early_results}). We believe this is due to the fact that AI data smoothes away all the hard case, therefore presenting the model with a cleaner signal in the initial stages.

\begin{table}[h]
    \small
    \centering
    \caption{\small \textbf{Correctly predicted GCD after 30, 60 and 90 million examples.} Dirty and correct datasets.}
    
    \begin{tabular}{c|cc|cc|cc}
    \toprule
        & \multicolumn{2}{c}{30M examples} & \multicolumn{2}{c}{60M examples} & \multicolumn{2}{c}{90M examples} \\
       Base  & AI & Correct & AI & Correct & AI& Correct  \\
       \midrule
       10 & 13 & 13 & 13 & 13 & 13 & 13 \\
       420 & 34 & 34 & 38 & 34 & 38 & 35\\
       1000 & 17 & 13 & 22 & 13 & 22 & 14\\
       2017 & 4 & 2 & 4 & 2 & 4 & 4\\
       2023 &  6 & 6 & 11 & 6  & 11 & 6\\
       4913 & 6 & 4 & 7 & 7 & 7 & 7\\
       \bottomrule
    \end{tabular}
    \label{tab:early_results}
\end{table}

This pattern changes after extensive training. Table~\ref{tab:late_results} compares performance of models trained on $300M$ and $1$ billion examples. For all bases $B$, models trained on $\mathcal C(B)$ learn new GCD as training proceeds, whereas models learned on $\mathcal D(B)$ never learn beyond their original performance.

\begin{table}[h]
    \small
    \centering
    \caption{\small \textbf{Correctly Predicted GCD after $300M$ and $1$ Billion Examples.} AI and correct datasets.}
    
    \begin{tabular}{c|cc|cc}
    \toprule
        & \multicolumn{2}{c}{300M examples} & \multicolumn{2}{c}{1B examples}  \\
       Base  & AI & Correct & AI & Correct  \\
       \midrule
       10 & 13 & 14 &13 & 31  \\
       420 & 38 & 38 &38 & 40  \\
       1000 & 22 & 25 &22 & 33  \\
       2017 & 4 & 6 &4 & 9  \\
       2023 & 16 & 16 &16 & 32 \\
       4913 & 17 & 16 &17 & 31 \\
       \bottomrule
    \end{tabular}
    \label{tab:late_results}
\end{table}

\begin{figure}
    \centering
    \includegraphics[width=0.3\linewidth]{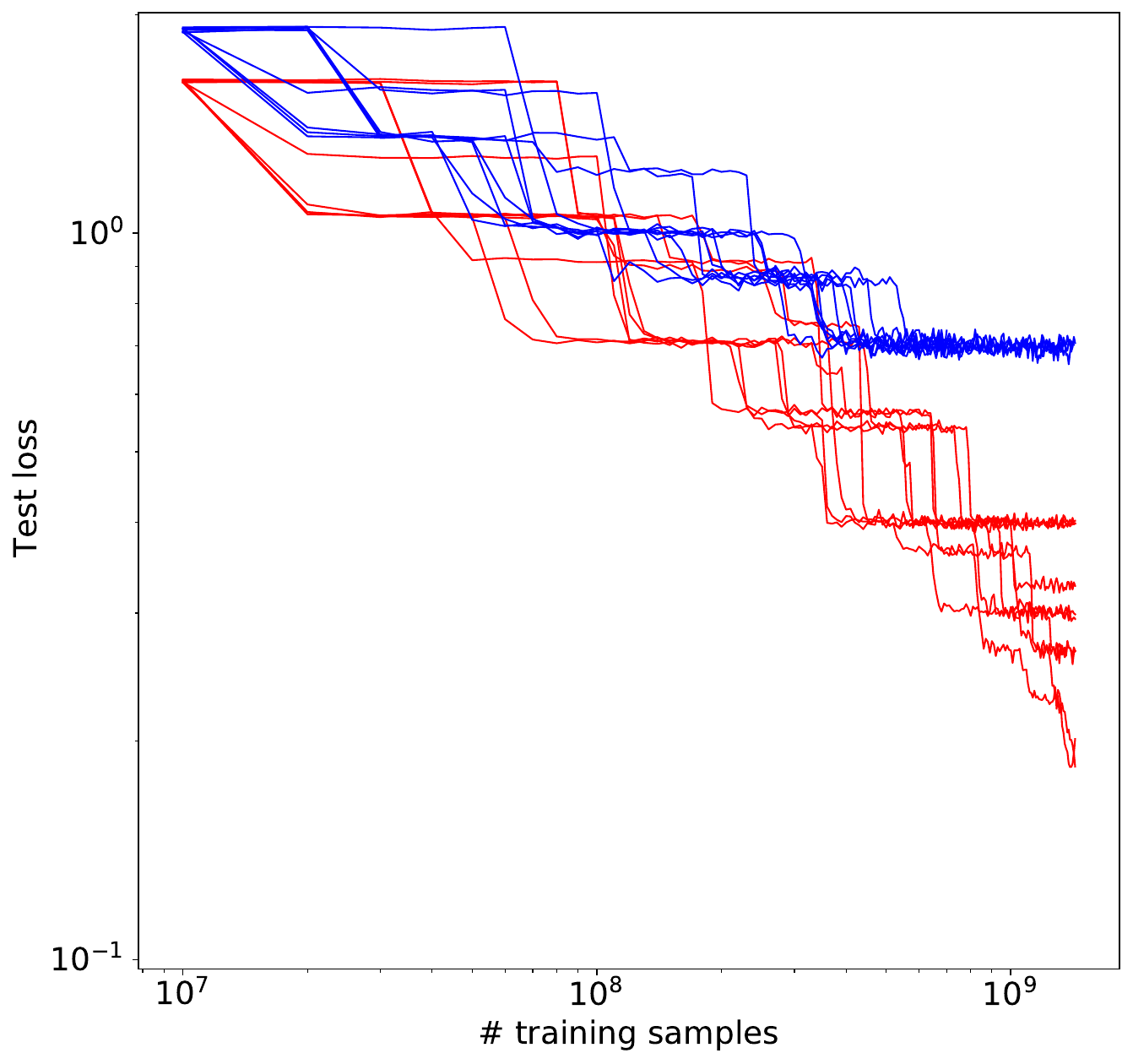}
    \includegraphics[width=0.3\linewidth]{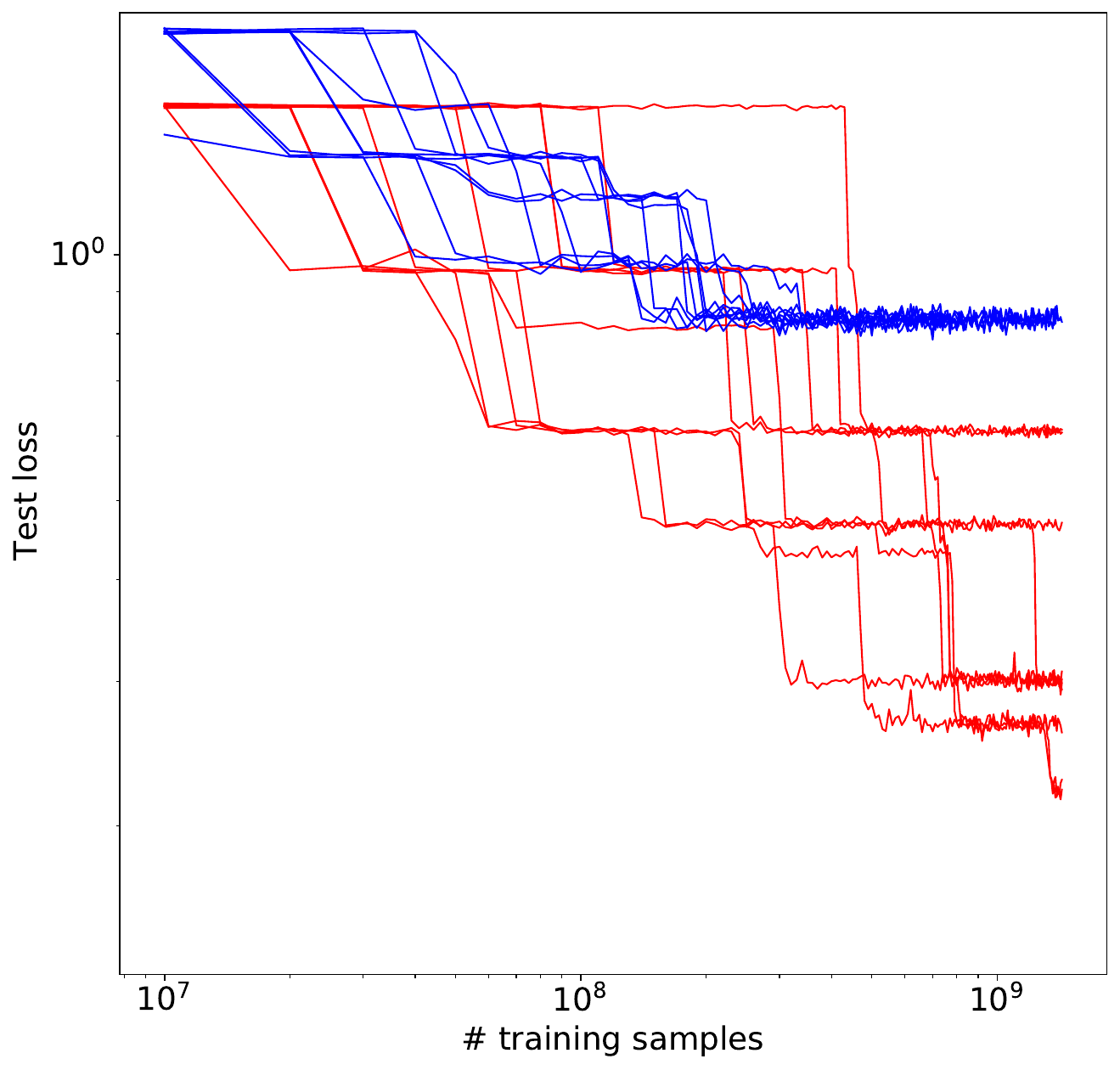}
     \includegraphics[width=0.3\linewidth]{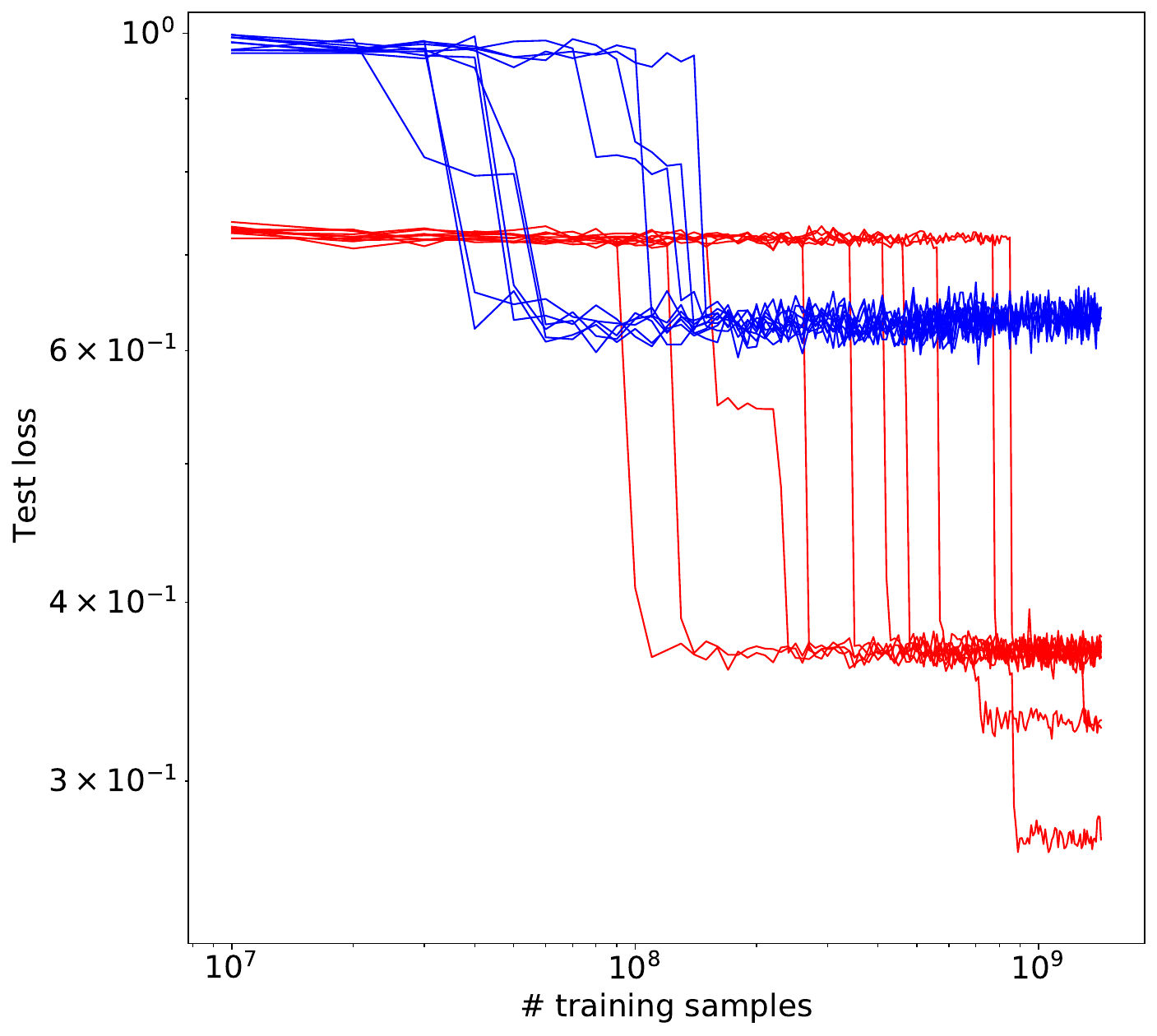}
    \caption{{\bf Test loss for GCD learning.} Test loss of 10 models trained on clean and generated data. From left to right: base 4913, 2023, 1000. Models trained on clean data (red) continune to learn (decreasing test loss) while models trained on AI generated data (blue) stops learning.} 
    \label{fig:all_base}
\end{figure}

\begin{figure}
    \centering
    \includegraphics[width=0.3\linewidth]{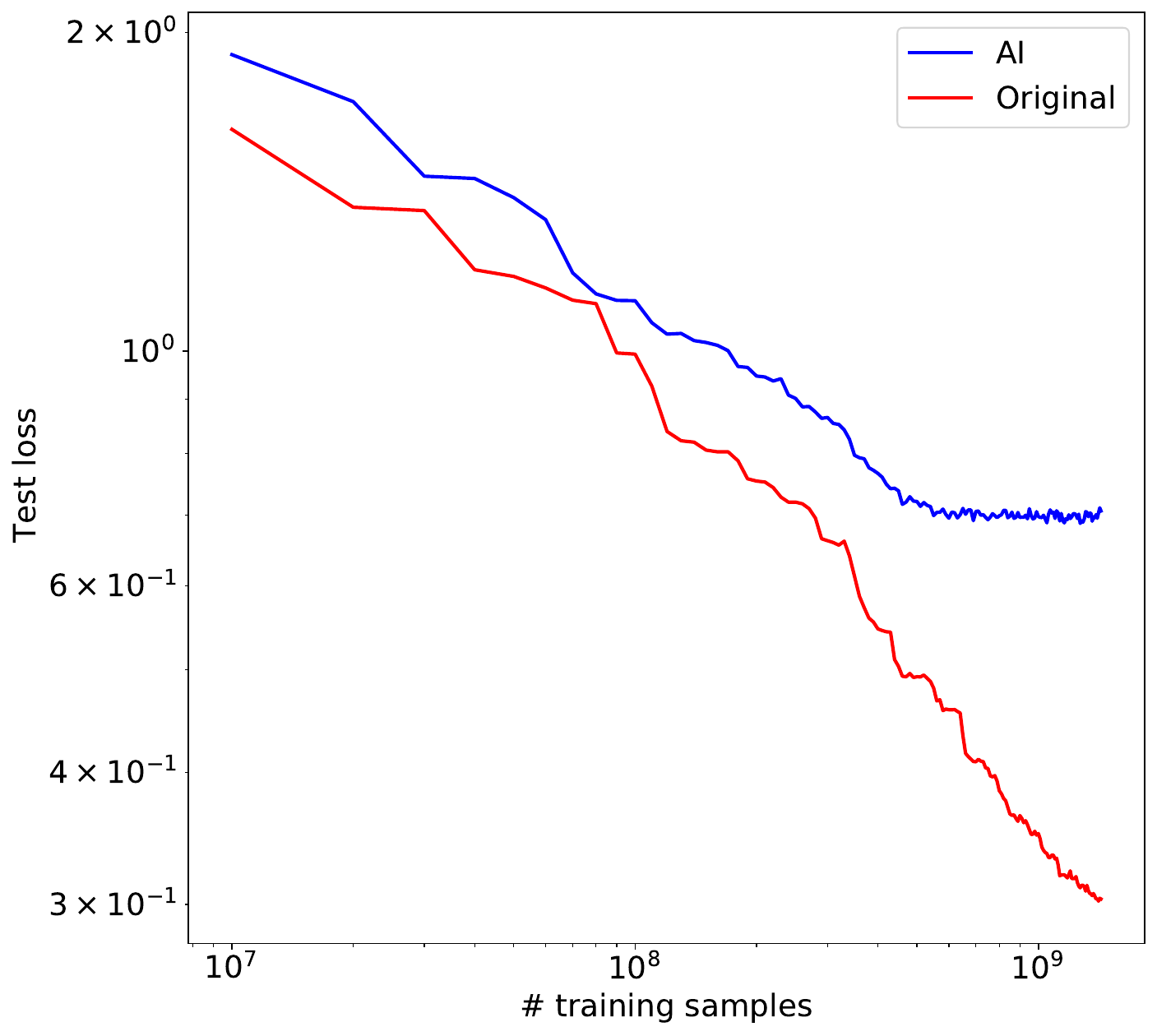}
      \includegraphics[width=0.3\linewidth]{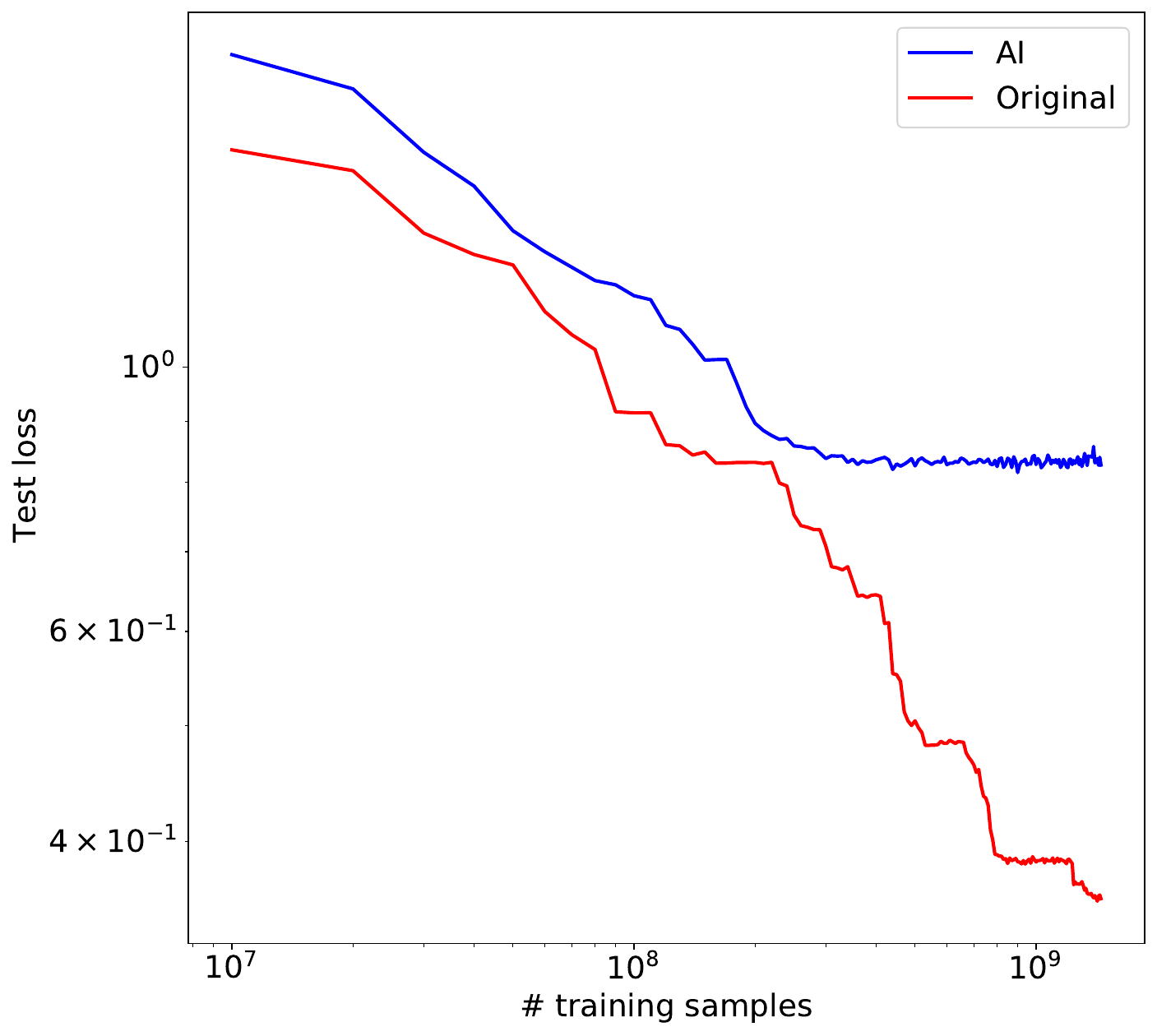}
       \includegraphics[width=0.3\linewidth]{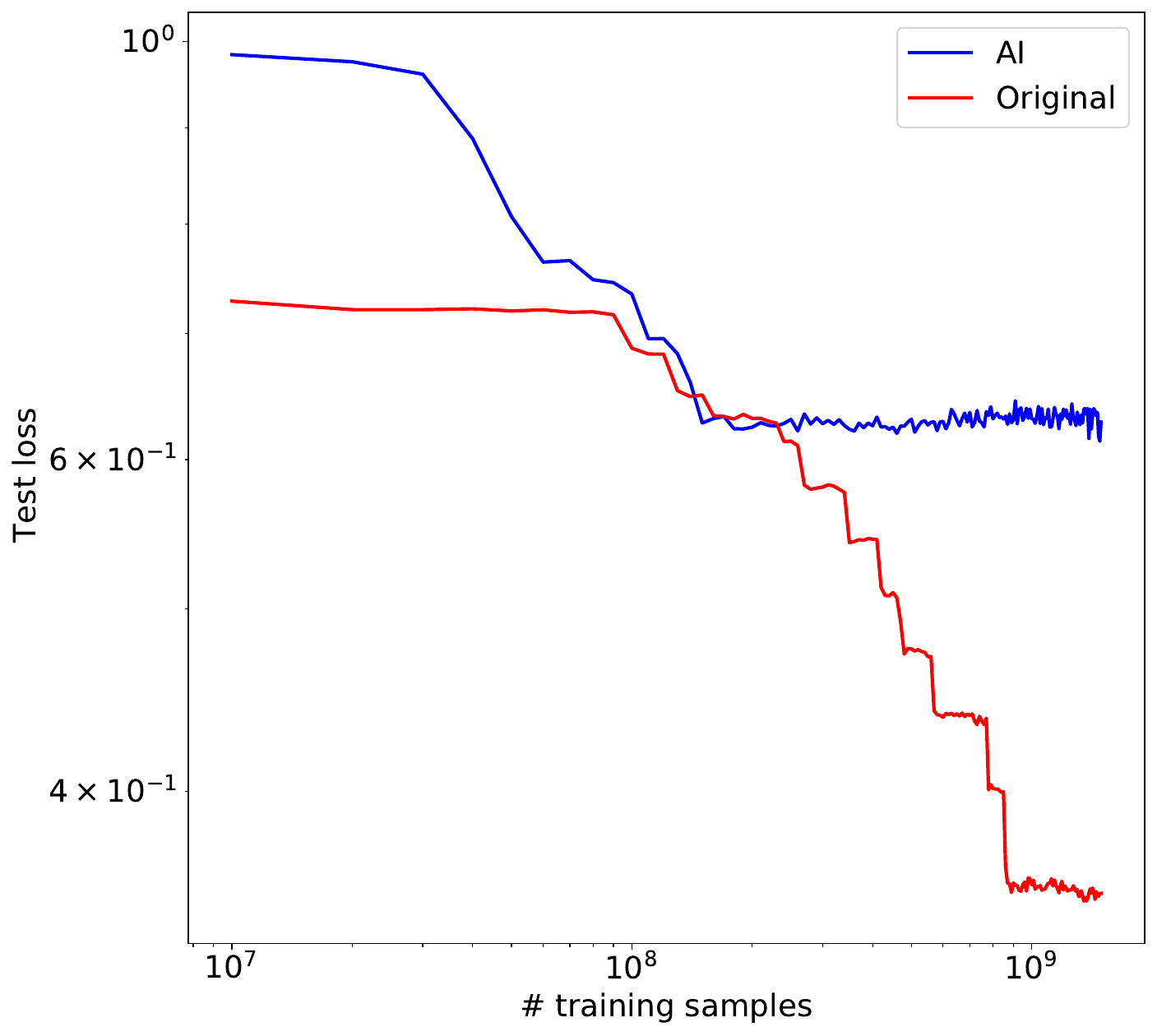}
    \caption{{\bf Average test loss for GCD learning.} Averaged over 10 models trained on clean and generated data. From left to right: base 4913, 2023, 1000.} 
    \label{fig:avg_base}
\end{figure}

\begin{figure}
    \centering
    \includegraphics[width=0.5\linewidth]{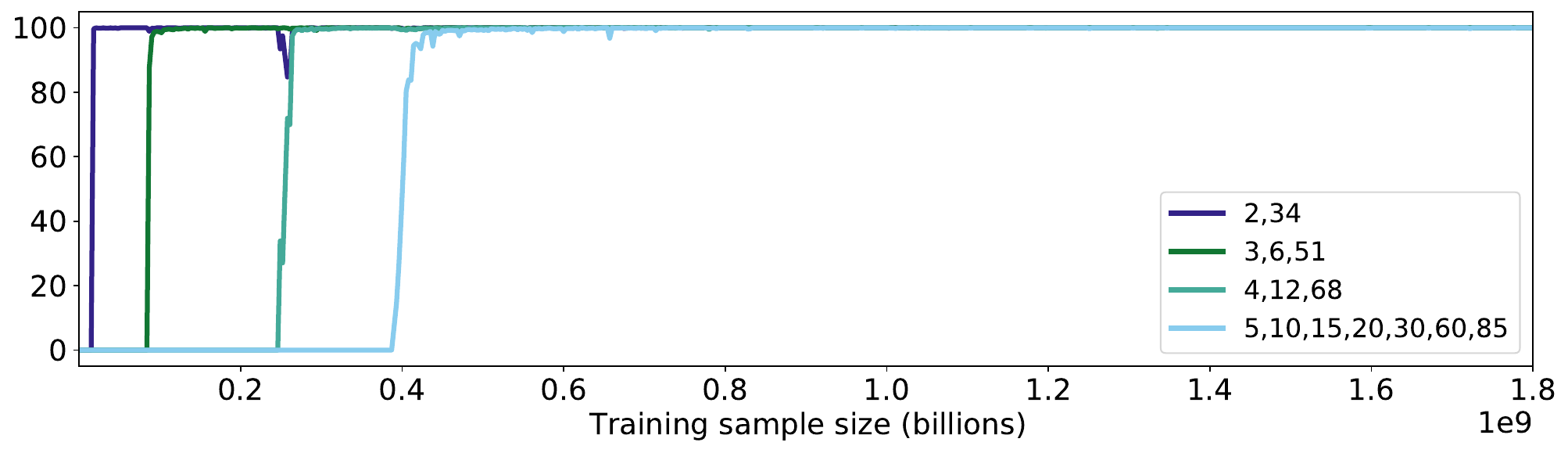}
      \includegraphics[width=0.5\linewidth]{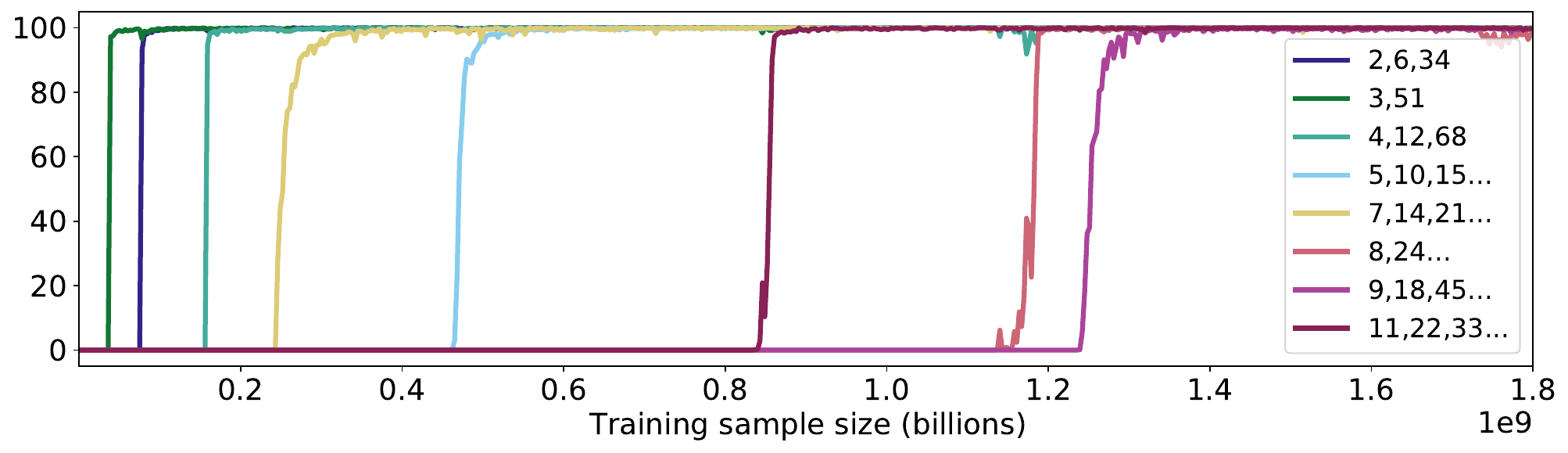}
    \caption{{\bf Emergence of skills (groups of GCDs learned together).}  Original (bottom) and AI-synthesized data (top). Base 4913. 1 model for clean/AI data, respectively.} 
    \label{fig:emerge}
\end{figure}

\begin{figure}
    \centering
    \includegraphics[width=0.5\linewidth]{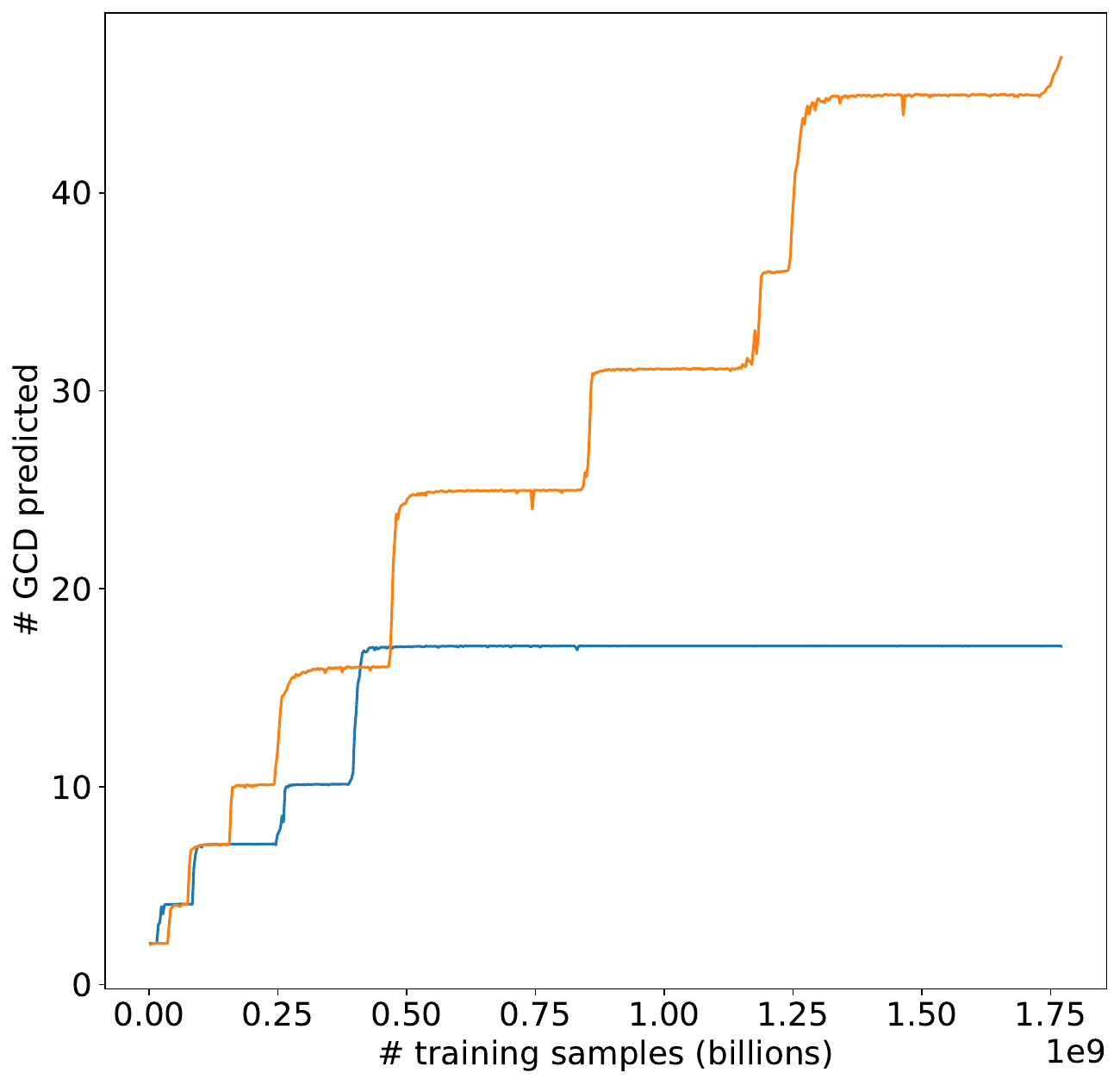}
    \caption{{\bf Learning the GCD.} Learning curve, base 4319. Orange: training on correct GCD. Blue: training on AI generated data.} 
    \label{fig:gcd_learningcurve}
\end{figure}

Figures \ref{fig:all_base} and \ref{fig:avg_base} show that we get the picture predicted by theory: the dirty model learns (until about 300M examples) and then stops learning (while the clean model continues) - its scaling law tapers off as predicted in Theorem \ref{thm:simple}. All the skills the clean model learns after this point are skills the model trained on synthesized data cannot learn (see Figure \ref{fig:emerge} showing when new learned groups of GCD emerge, and Figure \ref{fig:gcd_learningcurve} for the learning curve of two models, one trained on the original data, the other on AI data).

\paragraph{Mixing and Grokking}

\begin{figure}
    \centering
    \includegraphics[width=0.3\linewidth]{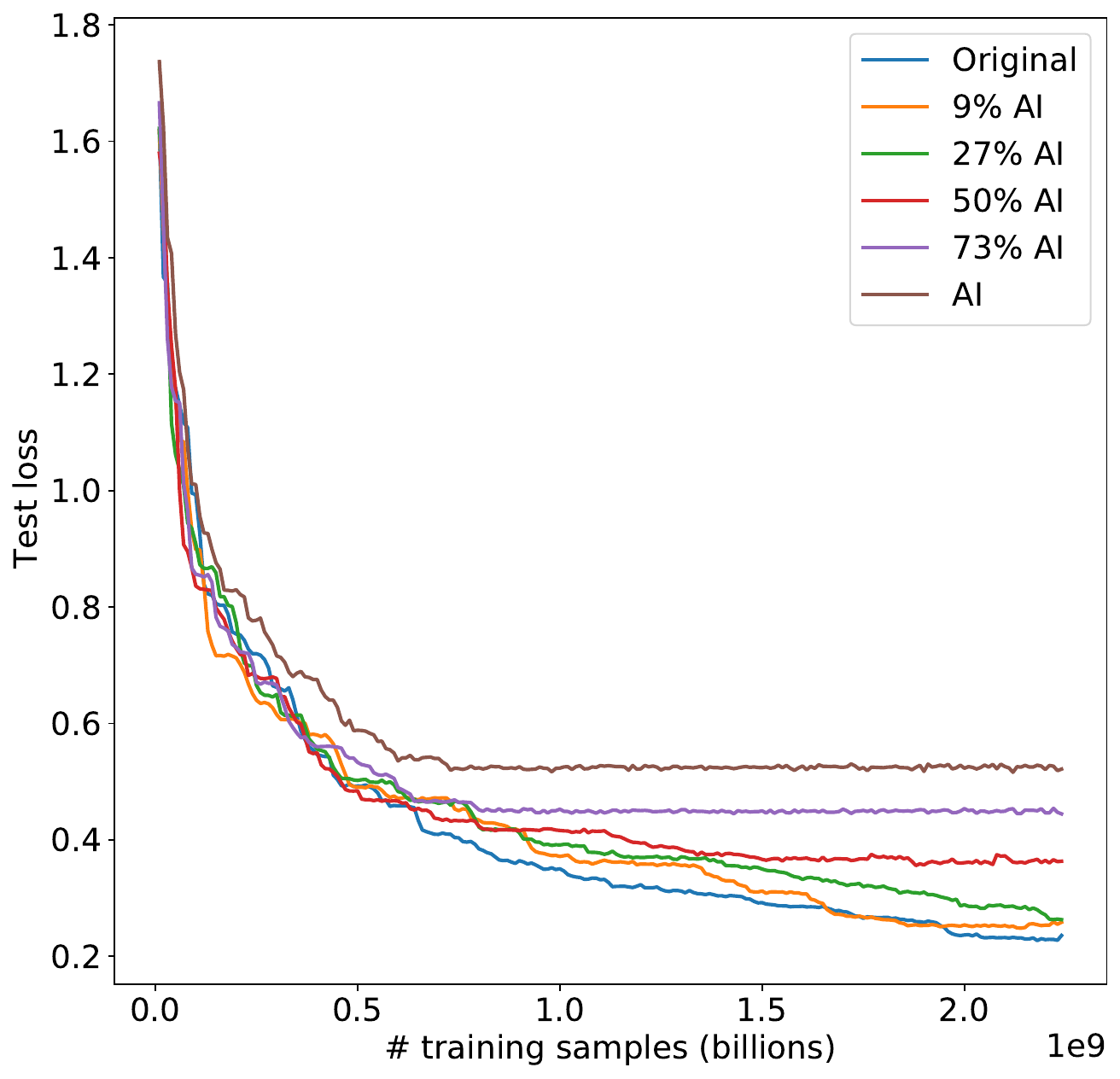}
     \includegraphics[width=0.3\linewidth]{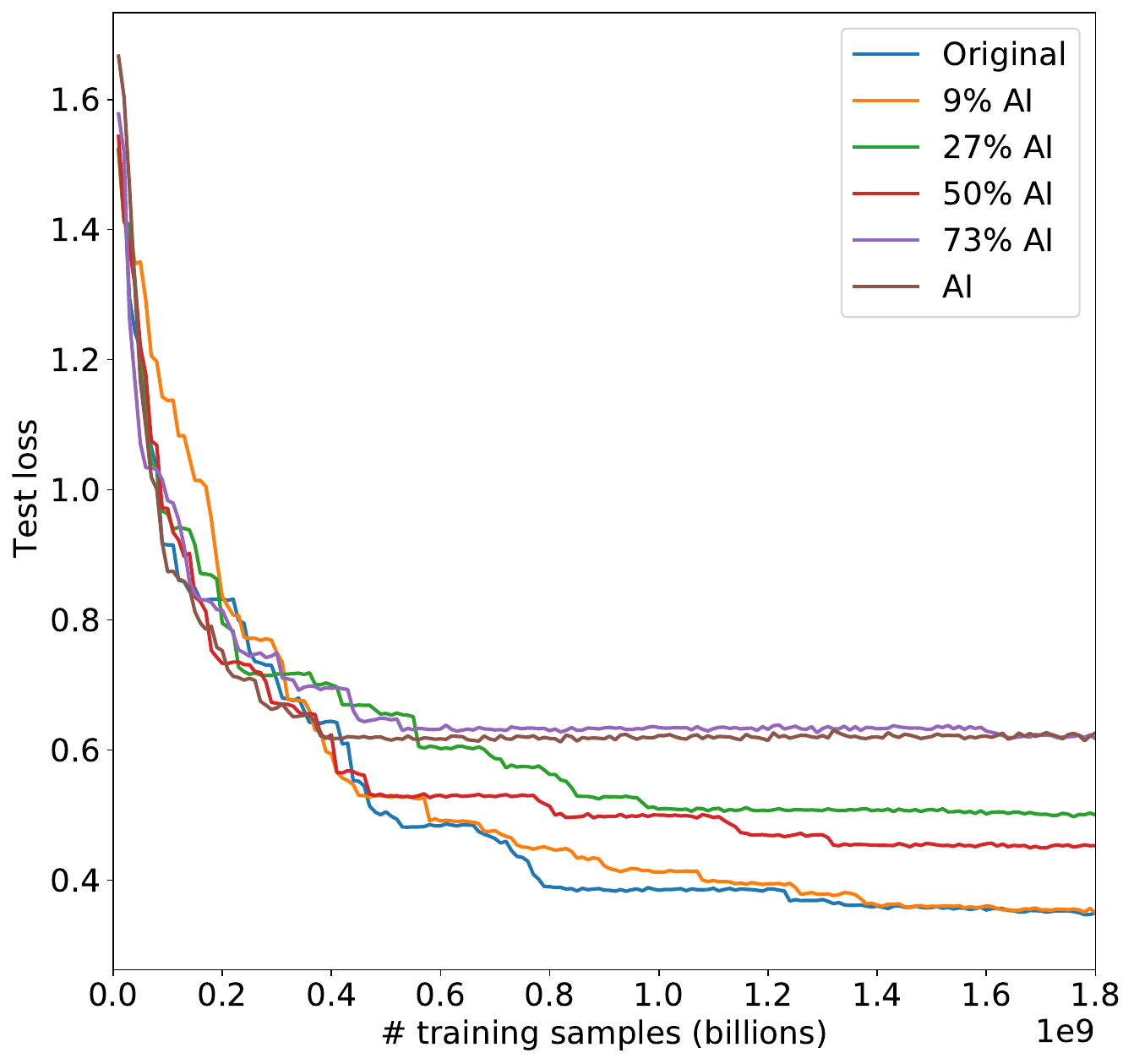}
     \includegraphics[width=0.3\linewidth]{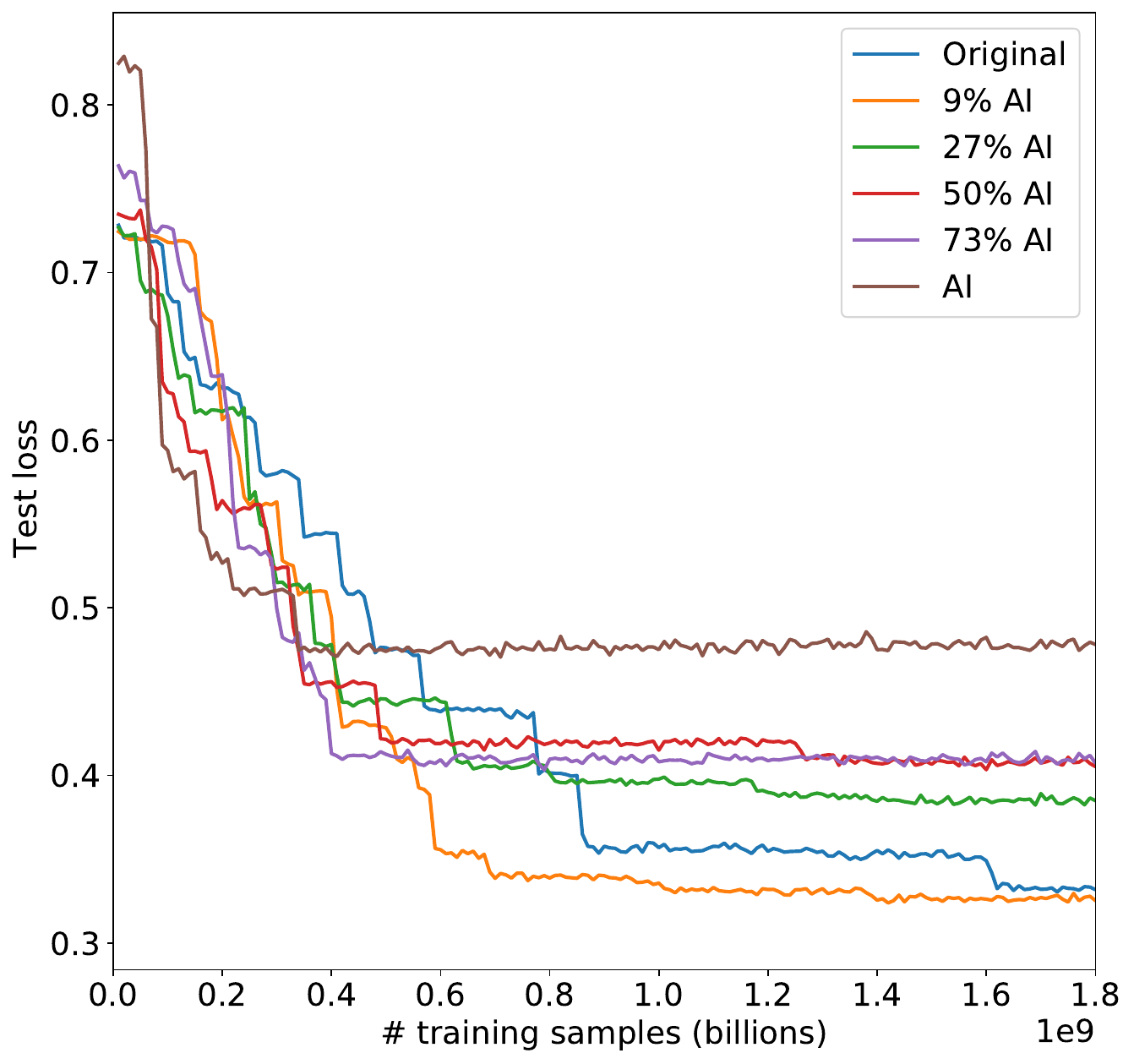}
    \caption{{\bf Grokking in GCD Learning on mixed data.} Error losses of models trained on mixtures of clean and AI generated GCD data. 10 models. From left to right: base 4913, 2023 and 1000.} 
    \label{fig:mix}
\end{figure}

We now proceed to train our model on randomly mixed clean and synthesized data for various mixture rates. 
We train with mixtures of clean and dirty data for mixture fractions of $9\%,27\%,50\%$ and $73\%$ of AI-generated data, for bases 1000, 2023 and 4913, to see the grokking effect. Figure \ref{fig:mix} illustrates the results. We can see that even for the average curves over the 10 seeds one can discern a grokking-like delayed learning for the mixtures with relatively small amounts of AI data. This effect can be studied 

The models used to generate the data were trained on about 300M examples, and correctly predict 22, 16 and 17 GCD below 100 for bases 1000, 2023 and 4913 respectively. We know (Table~\ref{tab:late_results}) that more training on AI-data data only will not improve those performances. On the other hand, we know that models trained on clean data will achieve larger performance. Specifically, out of 10 models trained on clean data, for base 1000, all 10 predict 23 GCD or more after 1.4B examples. The median number of examples needed for the models to predict 23 GCD or more is 465M. 
For base 2023, 7 models out of 10 predict 17 GCD or more after 2.1B examples. The median number of training samples after which the model bests a model trained on dirty data only is 530M.
Finally, for base 4913, 9 clean models out of  10 predict more than 18 GCD after 1.7B examples. The median number of samples is 1.1B.

When zooming in to when the mixture models learn to predict GCD that are ''unlearnable" with an AI-trained model, the grokking effect becomes more apparent. 

Table~\ref{tab:mixtures} summarizes by listing the time (\# of samples) when the mixture models finally learn a GCD that a purely AI-trained model cannot learn, and the delay (in millions samples) since the previous GCD was learned (see also Figure \ref{fig:emerge} to illustrate the comparison between the clean and the AI-trained model):

\begin{table}[h]
    \small
    \centering
    \caption{\small \textbf{Samples until Mixture Models Learn a GCD that AI-trained Models Cannot Learn.} * small number of experiments}
  
    \begin{tabular}{c|ccc|ccc|ccc}
    \toprule
        & \multicolumn{3}{c}{Base 1000} & \multicolumn{3}{c}{Base 2023} & \multicolumn{3}{c}{Base 4913}  \\
       mixture rate & successes & samples (M) & delay &  successes  & sample (M) & delay &successes  & samples (M)& delay \\
       \midrule
        0\% (clean) & 10/10 & 465 & 243 & 7/10 & 530 & 567 & 10/10 & 1180 & 520 \\
       9\% & 8/10 & 560 & 320 & 8/10 & 715 & 530 & 9/10 & 910 & 340\\
       27\% & 5/10 & 790 & 560 & 7/10 & 790 & 1220 & 10/10 & 1390 & 680\\
       50\% & 2/10$^*$ & 1310$^*$& 190$^*$ & 7/10 & 1140 & 1220 & 8/10 & 1280 & 1180\\
       73\% & 0 & - & -& 0 & - & -& 0 & -& - \\
       \bottomrule
    \end{tabular}
    \label{tab:mixtures}
\end{table}

The delay period increases with increasing fraction of AI data in the mix. Thus, Table \ref{tab:mixtures} clearly demonstrates the grokking effect of increasing plateau length with fraction of AI data, as predicted by our theory\footnote{We were constrained to stop the experiments at after about 3B samples for most, due to heavy use of compute resources. This probably explains why for the larger AI-mixtures only a few experiments could successfully find new GCDs - the other experiments where still in the pre-grokking phase when they were stopped.}.

\section{Details of Experiments with Llama2}\label{app:llama}

In the realm of large language models (LLMs), the prevailing approach involves a pretraining and finetuning paradigm. For instance, GPT-3 undergoes pretraining on approximately 45TB of text data from diverse sources. This extensive pretraining endows it with a robust capability for a variety of downstream tasks, employing methods such as zero-shot learning, few-shot learning, or finetuning. Our study evaluates the phenomenon of model collapse in scenarios close to the contemporary `synthetic data age.'

Utilizing one of the most advanced open-source models, {\tt Llama-2 7B}, our research investigates the effects on LLMs when they undergo finetuning\footnote{One can, in principle, replicate an experiment described here with training an LLM from scratch to demonstrate scaling law decay. We opted to not run such an experiment and instead focus on a more feasible finetuning setting, since just the language experiments described in the paper took weeks to run.
}  with data generated by other LLMs. To ensure the generation of high-quality data and to provide a relevant but not trivial downstream task, we employ the Wikitext-103 dataset. We segment this dataset into chunks of 128 tokens, between each with a stride of 64 tokens, resulting in approximately 2.2 million chunks. Denote this dataset as $\mathcal{D}_0$. The task for generation involves producing the final 32 tokens given the initial 96 tokens from each chunk in the original dataset. In the initial generation (0-th generation), we use the {\tt Llama-2 7B FT} model, which has been finetuned on $\mathcal{D}_0$, applying a generation loss that focuses solely on the cross-entropy loss of the final 32 tokens. We denote this initial model as $\mathcal{M}_0$, which demonstrates enhanced capacity for the generation task compared to the standard Llama-2 7B model. By querying $\mathcal{M}_0$ with the original 96 tokens from $\mathcal{D}_0$, we generate the dataset $\mathcal{D}_1$ and subsequently finetune {\tt Llama-2 7B} on this dataset to obtain $\mathcal{M}_1$. This process is sequentially repeated to generate $\mathcal{D}_{i}$ from $\mathcal{M}_{i-1}$ and obtain $\mathcal{M}_i$ through finetuning. By comparing the performance of various $\mathcal{M}$ models on the test set derived from Wikitext-103, also segmented into 128-token chunks, we aim to investigate the model collapse in LLMs.

To prevent information leakage across chunks, we restrict the training to only include the loss on the final 32 tokens for all generations. Consequently, the models are never trained on the first 96 tokens coming from the original corpus. The size of the 2.2 million chunks can provide sufficient data for finetuning while avoiding overfitting, given the capacity of {\tt Llama-2 7B}. Throughout the finetuning process, we maintain consistent settings using learning rate $5e^{-5}$ for LoRA, using Adam optimizer, dropout rate 0.1, trainable parameter fraction 0.062\%. To eliminate the possibility of model collapse due to insufficient sampling and to gain insights into scenarios where more AI-generated data is produced than the model has been trained (or finetuned) on, we consistently utilize a model trained on half the dataset for generating subsequent datasets.

\begin{figure}[htb]
    \centering
    \begin{minipage}{0.48\textwidth}
    \includegraphics[width=\linewidth]{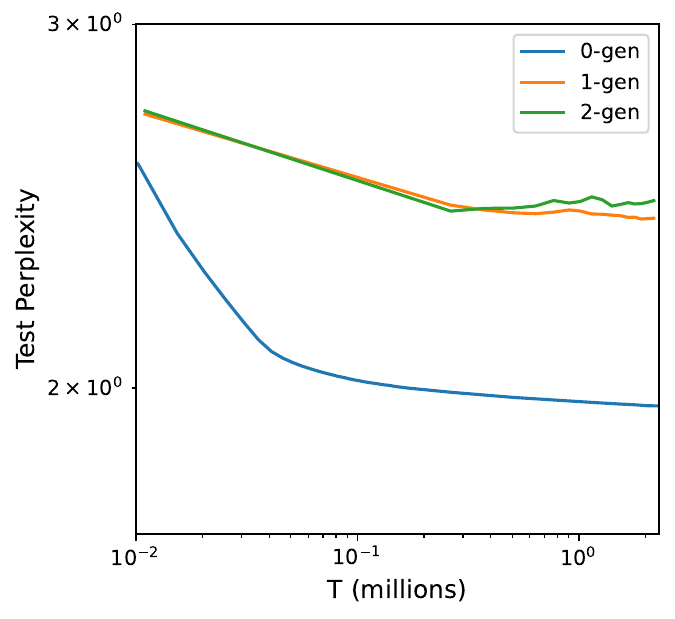}
    \caption{{\bf Llama Generated Data.} Llama2 finetuning when the loss for training and evaluation is the cross-entropy for all tokens in the chunks, including the prompt.}
    \label{fig:llama_YY}
    \end{minipage}
    \hfill
    \begin{minipage}{0.48\textwidth}
    \includegraphics[width=\linewidth]{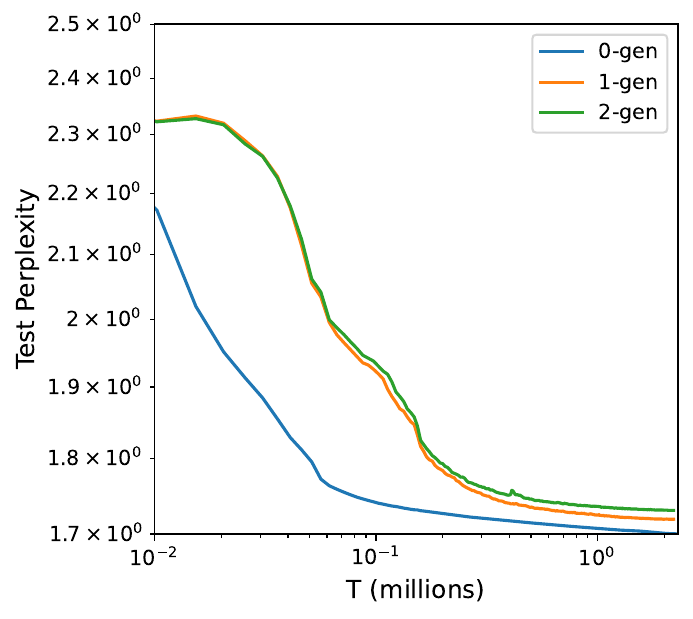}
    \caption{{\bf Mixing Llama Generated Data with Original Data.} Similar setting as Figure \ref{fig:experiments} left. Starting from gen 1, we mix the generated data with the original one with a ratio of 90 to 10. Top-$p^{inf}=$0.9 and temperature $\tau=$0.9.}
    \label{fig:llama_mix}
    \end{minipage}
\end{figure}

For completeness, we include Figure \ref{fig:llama_YY} with loss on the full chunks and Figure \ref{fig:llama_mix} that mix the generated data with original data. The mixing curve also aligns well with the grokking phenomenon predicted by theory.

\section{More Studies on Tail Cutting and Tail Narrowing Effects}\label{app:tail}

\begin{figure}
    \centering
    \includegraphics{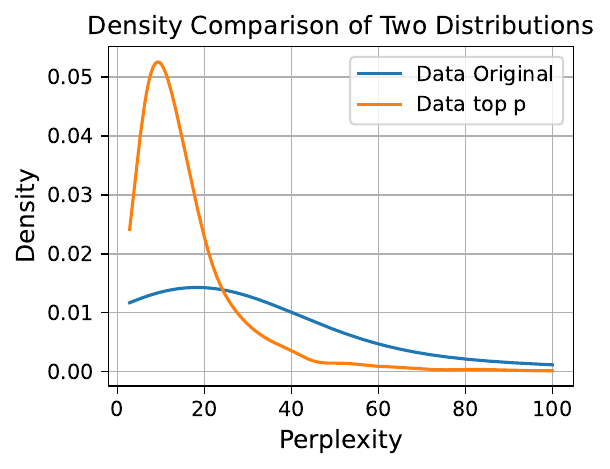}
    \caption{Sequential bigram data: top $p^{inf}$ = 0.95 leads to similar effect as tail narrowing. 1000 data with sequence length 100.} 
    \label{fig:cutoff=narrowing}
\end{figure}

Here, we illustrate how tail cutting in the next-token distribution can lead to tail-narrowing for metrics that take the entire sequence into account, like perplexity. Figure \ref{fig:cutoff=narrowing} illustrates this for the autoregressive bigram model. This effect is likely due to the combinatorial factors we obtain when considering an additive (or multiplicative) measure like perplexity.

\end{document}